\documentclass[fleqn,a4paper]{article}%

\usepackage{a4wide}
\usepackage{amsmath}
\usepackage{amsfonts}
\usepackage{amssymb}
\usepackage{graphicx}
\usepackage[numbers,square]{natbib}

\graphicspath{{graphics/}}

\usepackage{mathptmx}
\usepackage{helvet}
\usepackage{courier}

\usepackage{color}
\usepackage{url}

\newcommand{\Itau}[0]{\ensuremath{I^\tau}}
\newcommand{\Htau}[0]{\ensuremath{H^\tau}}
\newcommand{\lpzrobots}[0]{\textsc{LpzRobots}}
\newcommand{\Armband}[0]{\textsc{Armband}}
\newcommand{\Humanoid}[0]{\textsc{Humanoid}}

\newcommand{\infig}[1]{{\bf\textsf{#1}}}

\newcommand{\suppl}[1]{#1 on~\citep{videos:thispaper}}

\newcommand{\Fig}[1]{Figure~\ref{#1}}
\newcommand{\fig}[1]{Fig.~\ref{#1}}

\newcommand{\Eqn}[1]{Equation~\eqref{#1}}
\newcommand{\eqn}[1]{eq.~\eqref{#1}} 
\newcommand{\eqnp}[1]{(\eqn{#1})} 
\newcommand{\eqns}[2]{eqs.~\eqref{#1} and \eqref{#2}}
\newcommand{\eqnr}[2]{eqs.~\eqref{#1}-\eqref{#2}} 
\newcommand{\secA}[1]{section~\ref{#1}}
\renewcommand{\sec}[1]{section~\ref{#1}}

\newcommand{\period}{} 
\newcommand{\komma}{}  

\newcommand{\ie}{i.\,e.~}
\newcommand{\eg}{e.\,g.~}
\newcommand{\rhs}{r.\,h.\,s.~}
\newcommand{\dint}{\,\mathrm{d}} 

\newcommand{\Real}{\ensuremath{\mathbb R}}        
\newcommand{\T}{\ensuremath{\top}}                

\newcommand{\Tr}{\ensuremath{\mathrm{Tr}}}        

\usepackage[pdftex,colorlinks=false]{hyperref}

\newcommand{\Todo}[1]%
{{\bf\framebox{TODO: #1}}}

\begin{document}

\title{Information driven self-organization of complex robotic behaviors}
\author{Georg Martius$^{1}$, Ralf Der$^{1}$, Nihat Ay$^{1,2}$\\$^{1}$Max Planck Institute for Mathematics, Leipzig, Germany\\$^{2}$Santa Fe Institute, Santa Fe, USA\\
  {\small \tt \{martius|ralfder|nay\}@mis.mpg.de}
}
\maketitle
\date

\begin{abstract}
Information theory is a powerful tool to express principles to drive autonomous systems
 because it is domain invariant and allows for an intuitive interpretation.
This paper studies the use of the predictive information (PI),
also called excess entropy or effective measure
complexity, of the sensorimotor process as a driving force to generate behavior.
We study nonlinear and nonstationary systems and introduce the
time-local predicting information (TiPI) which allows us to derive exact results
 together with explicit update rules for the parameters of the
 controller in the dynamical systems framework.
In this way the information principle, formulated at the
level of behavior, is translated to the dynamics of the synapses.
We underpin our results with a number of case studies with high-dimensional robotic
 systems.
We show the spontaneous cooperativity in a complex physical system with decentralized control.
Moreover, a jointly controlled humanoid robot develops a high behavioral variety depending on its
 physics and the environment it is dynamically embedded into.
The behavior can be decomposed into a succession of low-dimensional modes that
 increasingly explore the behavior space.
This is a promising way to avoid the curse of dimensionality
 which hinders learning systems to scale well.
\end{abstract}

\section{Introduction}

Autonomy is a puzzling phenomenon in nature and a major challenge in the world
of artifacts. A key feature of autonomy in both natural and artificial systems
is seen in the ability for independent exploration~\citep{Boden08}.
In animals and humans, the
ability to modify its own pattern of activity is not only an indispensable
trait for adaptation and survival in new situations, it also provides a
learning system with novel information for improving its cognitive capabilities,
and it is essential for development.
Efficient exploration in high-dimensional spaces is a major
 challenge in building learning systems.
The famous exploration-exploitation trade-off was extensively studied
 in the area of reinforcement learning~\citep{suttonbarto98book}.
In a Bayesian formulation this trade-off can be
optimally solved~\citep{Duff:optimal-learning}, however it is computationally intractable.
A more conceptual solution is to provide the agent with an intrinsic motivation~\citep{Schmidhuber91:CuriousControl,Singh:2010}
 for focusing on certain things and thus constraining the exploration to a smaller space.
To approach this problem in a more fundamental way we consider
 mechanisms for goal-free exploration of the  dynamical
 properties of a physical system, \eg a robot.
If the exploration is rooted in the agent in a self-determined way, \ie as a
deterministic function of internal state variables and not via a pseudo-random
 generator it has the chance to escape the curse of dimensionality.
Why? Because specific features of the system such as constrains and other embodiment effects
 can be exploited to reduce the search space.
Thus an exploration strategy taking the particular body and environment
 into account is vital for building efficient learning algorithms for high-dimensional robotic systems.
But how can goal-free exploration be useful to actually pursue goals?
We show that a variety of coordinated sensorimotor patterns are formed
 that may be used to quickly construct more complex behaviors using a second level of learning.
It may also be used more directly in combination with reinforcement learning where the typical random
 exploration is substituted or augmented by the goal-free exploration leading presumably to a large speedup.

The solution for such a general problem needs a core paradigm in order to be relevant for
a large class of systems. In recent years, information theory has come into
the focus of researchers interested in a number of related issues ranging from
quantifying and better understanding autonomous systems~\citep{BertschingerOAJ08,lungarella06:InfoFlowInSML,Friston96,Sporns02classesof,garofalo09,williamsbeer10:infordyn,Hoffmann2012}
to questions of spontaneity in biology and technical
systems~\citep{Brembs11} to the self-organization of robot behavior~\citep{AyDerBernigauProkopenko2012,ZahediAyDer2010:HigherCoordination}.

A systematic approach requires both a convenient definition of the information
measure and a robust, real time algorithm for the maximization of that
measure. This paper studies in detail the use of the predictive
information (PI) of a robot's sensorimotor process. The predictive information
of a process quantifies the total information of past experience that can be
used for predicting future events. Technically, it is defined as the mutual
information between the past and the future of the time series. It has
been argued~\citep{bialek01} that predictive information, also termed excess
entropy~\citep{Crutchfield89} and effective measure complexity
\citep{Grassberger86d}, is the most natural complexity measure for time series.
By definition, predictive information of the sensor process is high
 if the robot manages to produce a stream of sensor values
 with high information content (in the Shannon sense)
 by using actions that lead to predictable consequences.
A robot maximizing PI therefore is expected to show a high
variety of behavior without becoming chaotic or purely random. In this working
regime, somewhere between order and chaos, the robot will explore its
behavioral spectrum in a self-determined way in the sense discussed above.

This paper studies the control of robots by simple neural networks
 whose parameters (synaptic strengths and threshold values) are adapted on-line
 to maximize (a modified) PI of the sensor process.
These rules define a mechanism for behavioral variability as a deterministic
 function formulated at the synaptic level.
For linear systems a number of features of the PI maximization
method have been demonstrated~\citep{AyDerBernigauProkopenko2012}. In
particular, it could be shown that the principle makes the system to
explore its behavior space in a systematic manner. In a specific case, the
PI maximization caused the controller of a stochastic oscillator system to
sweep through the space of available frequencies. More importantly, if the
world is hosting a latent oscillation, the controller will learn
 by PI maximization to go into resonance with this
inherent mode of the world. This is encouraging, since maximizing the PI
means (at least in this simple example) to recognize and amplify the latent
modes of the robotic system.

The present paper is devoted to the extension of the above mentioned method
to nonlinear systems
 with nonstationary dynamics.
This leads to a number of novel elements in the present approach.
Commonly information theoretic measures are optimized in the
stationary state.
This is not adequate for a robot in a
self-determined process of behavioral development. This paper develops
a more appropriate measure for this purpose called the
 time-local predictive information (TiPI)
 for general nonstationary processes by using a specific
 windowing technique and conditioning.
Moreover, the application of information theoretic
measures in robotics is often restricted to the case of a finite state-action
space with discrete actions and sensor values. Also these restrictions
 are overcome in this paper so that it can be used
immediately in physical robots with high dimensional state-action
space. This will be demonstrated by examples with two robots in a
physically realistic simulation. The approach is seen to work from scratch,
\ie without any knowledge about the robot, so that everything has to be
inferred from the sensor values alone.
In contrast to the linear case the nonlinearities
and the nonstationarity introduce a number of new phenomena,
 for instance the self-switching dynamics in a simple hysteresis system and
 the spontaneous cooperation of physically coupled systems.
In high-dimensional systems we observe behavioral patterns of
 reduced dimensionality that are dependent on the body and the environment of the robot.

\subsection{Relation to other work}

Finding general mechanisms that help robots and other systems to more
autonomy, is the topic of intensive recent research. The approaches are widely
scattered and follow many different routes so that we give in the following
just a few examples.

\subsubsection{Information theoretic measures}

Information theory has been used recently in a number of approaches in
robotics in order (i) to understand how input information is structured by the
behavior~\citep{Lungarella2005,lungarella06:InfoFlowInSML} and (ii) to quantify the nature of information flows inside the brain~\citep{Friston96,Sporns02classesof,garofalo09} and
in behaving robots~\citep{Hoffmann2012,williamsbeer10:infordyn}.
An interesting information measure is the empowerment,
 quantifying the amount of Shannon information that an agent can
``inject into'' its sensor through the
environment, affecting future actions and future perceptions. Recently,
empowerment has been demonstrated to be a viable objective for the
self-determined development of behavior in the pole balancer problem
and other agents in continuous domains~\citep{PolaniJung11}.

Driving exploration by maximizing PI can also be considered as an alternative to
the principle of homeokinesis as introduced in~\citet{DerLieb02,Der01} that has
been applied successfully to a large number of complex robotic systems,
see~\citet{DerHesseMartius05,derfeeling05,derspherical06,dermartius:babble06}
and the recent book~\citet{DerMartius11}. Moreover,
this principle has also been extended to form a basis for a guided
self-organization of behavior~\citep{martius:guidedselforg07,DerMartius11}.

\subsubsection{Intrinsic motivation}

As mentioned above, the self-determined and self-directed exploration for embodied autonomous agents
is closely related to many recent efforts to equip the robot with a motivation
system producing internal reward signals for reinforcement learning in pre-specified tasks.
Pioneering work has been done by Schmidhuber using the
prediction progress as a reward signal in order to make the robot curious for new
experiences~\citep{Schmidhuber116542,Storck1995Reinforcement-Driven-Information,Schmidhuber2009Driven-by-Compression}.
Related ideas have been put
forward in the so called play ground experiment~\citep{Kaplan2004Maximizing-Learning-Progress:,Oudeyer2007Intrinsic-Motivation-Systems}.
There have been also a few proposals to autonomously form a hierarchy of competencies
 using the prediction error of skill models~\citep{Barto04intrinsicallymotivated}
 or more abstractly to balance skills and challenges~\citep{Steels2004The-Autotelic-Principle}.
Predictive information can also be used as an intrinsic motivation
  in reinforcement learning~\citep{ZahediMartiusAy12:gso5}
or additional fitness in evolutionary robotics~\citep{prokopenko:sab06}.

\subsubsection{Embodiment}
The past two decades in robotics have seen the emergence of a new trend of control
in robotics which is rooted more deeply in the dynamical systems approach to
robotics using continuous sensor and action variables.
This approach yields more natural movements of the robots and allows to exploit embodiment effects in an effective way, see~\citet{PfeiferBongar2006:BodyShapesThink,Pfeifer07} for an excellent survey.
The approach described in the present paper is tightly coupled to the ideas
 of exploiting the embodiment, since the development of behavioral modes
 is entire dependent on the dynamical coupling of the body, brain, and its environment.

\subsubsection{Spontaneity}
We would like to briefly discuss the implications of using
 a self-determined\footnote{Self-determined is understood here has ``only based its own
 internal laws''}
and deterministic mechanism of exploration to the understanding of
 variability in animal behavior. 
In the animal kingdom, there is
increasing evidence showing that animals from invertebrates to fish,
birds, and mammals are equipped with a
surprising degree of variety in response to external
stimulation~\citep{bekoff98:animal_play,glickman66:curios,stoewe06:novel_exploration_raven,berlyne66:curios_exploration}.
So far, it is not clear how this behavioral variability is created. Ideas
cover the whole range from the quantum effects~\citep{Koch09:freewill}
 (pure and inexorable randomness)
to thermal fluctuations at the molecular level to the assumption
of pure spontaneity~\citep{Maye07:order_in_spontaneous},
 rooting the variability in the existence of intrinsic, purely deterministic processes.

This paper shows that a pure spontaneity is enough to produce behavioral
 variations, and as in animals, their exact source appears ``indecipherable''
 from an observer point of view.
If the variation of behavior in animals is produced in a similar way,
 this would bring new insights into the free will conundrum~\citep{Brembs11}.

\section{Methods}

We start with the general expressions for the predictive information (PI) and introduce
 a derived quantity called time-local predictive information (TiPI) more
suitable for the intended treatment of nonstationary systems.
Based on the specific choice of the time windows we derive estimates of the TiPI for general stochastic dynamical systems and give explicit expressions for the special case of a Gaussian noise.
The explicit  expressions are used for the derivation of
the parameter dynamics of the controller (exploration dynamics) obtained by gradient ascending the TiPI.
Besides giving the exploration dynamics as a batch rule we also derive,
 in the sense of a stochastic gradient rule, the one-shot gradient.
The resulting combined dynamics (system plus exploration dynamics)
 is a deterministic dynamical system, where the self-exploration of the system
 becomes a part of the strategy.
These general results are then applied to the case of the sensorimotor loop
 and we discuss their Hebbian nature.

\subsection{Predictive information}

The PI of a time discrete process $\left\{  S_{t}\right\}
_{t=a}^b$ with values in $\Real^n$ is defined~\citep{bialek01} as the mutual information
between the past and the
future, relative to some instant of time $a \le t_0 < b$
\begin{equation}
I\left( S_{\text{future }};S_{\text{past}}\right)  =\left\langle \ln
\frac{p\left(  s_{\text{future }},s_{\text{past}}\right)  }{p\left(
s_{\text{past}}\right)  p\left(  s_{\text{future}}\right)  }\right\rangle
=H\left(  S_{\text{future }}\right)  -H\left(  S_{\text{future }}|S_{\text{past}}\right)  \label{eqn:PredInfo:Definition}%
\end{equation}
where the averaging is over the joint probability density distribution
$p\left(  s_{\text{past}},s_{\text{future }}\right)$ with $\text{past}:=\{a,\dots,t_0\}$ and $\text{future}:=\{t_0+1,\dots,b\}$.
In more detail, we use
the (differential) entropy $H\left(  S\right)  $ of a random variable $S$
given by
\[
H\left(  S\right)  =- \int p\left(  s\right)  \ln p\left(  s\right)  \dint s
\]
where $p\left(  s\right)  $ is the probability density distribution of the
random variable $S$. The conditional entropy $H\left(  S_t|S_{t-1}\right)  $
is defined accordingly
\[
H\left(  S_t|S_{t-1}\right)  =-\int\int p\left(  s_t|s_{t-1}\right)  \ln
p\left(  s_t|s_{t-1}\right)  \dint s_{t} \, p(s_{t-1}) \dint s_{t-1}
\]
$p\left(  s_t|s_{t-1}\right)  $ being the conditional probability density
distribution of $s_t$ given $s_{t-1}$. As is well known, in the case of
continuous variables, the individual entropy components $H\left(
S_{\text{future }}\right)  $, $H\left(  S_{\text{future }}|S_{\text{past}%
}\right)  $ may well be negative whereas the PI is always positive and will
exist even in cases where the individual entropies diverge. This is a very
favorable property deriving from the explicit scale invariance of the PI
\citep{AyDerBernigauProkopenko2012}.

The usefulness of the PI for the development of explorative behaviors of
autonomous robots has been discussed earlier\footnote{In experiments with a
coupled chain of robots~\citep{DerGuettlerAy08:predinf} it was
observed that the PI\ of just a single sensor, one of the wheel counters of an
individual robot, already yields essential information on the behavior of the
robot chain. The PI turned out to be maximal if the individual robots managed
to cooperate so that the chain as a whole could navigate effectively. This is
remarkable in that a one-dimensional sensor process can already give essential
information on the behavior of a very complex physical object under real world
conditions. These results give us some encouragement to study the role of PI
and other information measures for specific sensor processes as is done in the
present paper.}, see
\citet{ay08:predinf_explore_behavior,DerGuettlerAy08:predinf,AyDerBernigauProkopenko2012}%
. This paper continues these investigations for the case of more general
situations than those discussed before. In order to do so, we have to
introduce some specifications necessary for the development of a
versatile and stable algorithm realizing the increase of PI in the sensor
process at least approximately.

Let us start with simplifying \eqn{eqn:PredInfo:Definition}. If
$\left\{  S_{t}\right\}  _{t=a}^{b}$ is a Markov process, see
\citet{ay08:predinf_explore_behavior}, the PI is given by the mutual
information (MI) between two successive time steps, \ie  instead of
\eqn{eqn:PredInfo:Definition} we have
\begin{equation}
I\left(  S_{t};S_{t-1}\right)  =\left\langle \ln\frac{p\left(  s_{t}%
,s_{t-1}\right)  }{p\left(  s_{t}\right)  p\left(  s_{t-1}\right)
}\right\rangle =H\left(  S_{t}\right)  -H\left(  S_{t}|S_{t-1}\right)
\label{eqn:PredInfo:Markov}%
\end{equation}
the averaging being done over the joint probability density $p\left(
s_{t},s_{t-1}\right)  $. Actually, any realistic sensor process will
 only be in exceptional cases purely Markovian. However, we can use the
simplified expression \eqref{eqn:PredInfo:Markov}---let us call it the one-step
PI---also for general sensor processes taking it as the \textbf{definition} of
the objective function driving the autonomous exploration dynamics to
be derived.

\subsubsection{Nonstationarity and time-local predictive information (TiPI)}

Most applications done so far were striving for the evaluation of the PI
 in a stationary state of the system.
With our robotic applications, this is neither necessary nor adequate.
The robot is to develop a variety of behavioral modes ideally in a open-ended fashion,
 which will certainly not lead to a stationary distribution of sensor values.
The PI would change on the timescale of the behavior.
How can one obtain in this case the probability distributions of $p(s_t)$?
The solution we suggest is to introduce a conditioning on an initial state
in a moving time window and thus obtain the distributions from our local model as introduced below.
More formally, let us consider the following
setting. Let $t$ be the current instant of time and $\tau$ be the length of
 a time window $\tau$ steps into the past. We study the process in that window
with a fixed starting state $s_{t-\tau}$ so that all
distributions in \eqn{eqn:PredInfo:Markov} are conditioned on state
$s_{t-\tau}$. For instance, instead of $p\left(  s_{t}\right)  $ in
\eqn{eqn:PredInfo:Markov}, we have to use\footnote{In the Markovian case this
boils down to
\[
p\left(  s_{t}|s_{t-\tau}\right)  =\int\cdots\int p\left(  s_{t}%
|s_{t-1}\right)  \cdots p\left(  s_{t-\tau+1}|s_{t-\tau}\right)
ds_{t-1}\cdots \dint s_{t-\tau+1}%
\]
}
\begin{equation}
p\left(  s_{t}|s_{t-\tau}\right)  =\int\cdots\int p\left(  s_{t}%
,\ldots,s_{t-\tau+1}|s_{t-\tau}\right)  \dint s_{t-1}\cdots \dint s_{t-\tau+1}
\label{eqn:PredInfo:Markov:pvons}%
\end{equation}
and the related expression for $p\left(  s_{t},s_{t-1}|s_{t-\tau}\right)  $,
where $p\left(  s_{t},\ldots,s_{t-\tau+1}|s_{t-\tau}\right)$ is the joint
probability distribution for the process in the time window, conditioned on
$s_{t-\tau}$.
As to notation, the conditional probabilities depend explicitly
on time so that $p\left(  s_{t^{\prime}}|s_{t^{\prime}-1}\right)  $ is
different from $p\left(  s_{t^{\prime\prime}}|s_{t^{\prime\prime}-1}\right)  $
in general if $t^{\prime}\neq t^{\prime\prime}$, with equality only in the
stationary state. As a result we obtain the new quantity, written in a short-hand notation as
\begin{equation}
\Itau\left(S_t;S_{t-1}\right):= I\left(S_t;S_{t-1} | S_{t-\tau} = s_{t-\tau}\right)\label{eqn:TiPI}
\end{equation}
which we call \emph{time-local predictive information} (TiPI).
Note the difference to the conditional mutual information where an averaging over $s_{t-\tau}$ would take place. Analogously we define the time local entropy as
\begin{equation}
\Htau\left(S_t\right):= H\left(S_t| S_{t-\tau} = s_{t-\tau}\right)\period\label{eqn:Htau}
\end{equation}

\subsection{Estimating the TiPI}

To evaluate the TiPI only the kernels have to be known which
can be sampled by the agent on the basis of the measured sensor values.
However, in order to get explicit update rules driving the increase of the
TiPI, these kernels have to be known as a function of the parameters of the
system, in particular those of the controller. This can be done by learning
the kernels as a function of the parameters. A related approach, followed in
this paper, is to learn a model of the time series, i.e. learning a function
$\psi:\Real^{n}\rightarrow\Real^{n}$ acting as a time series
predictor $S_{t}=\psi\left(  S_{t-1}\right)  +\Xi$ with realization
\begin{equation}
s_{t}=\psi\left(  s_{t-1}\right)  +\xi_{t} \label{eqn:estimat:PI:psi}%
\end{equation}
for any time $t$, $\xi_t$ being the prediction error, also called the noise in
the following. $\psi$ can be realized for instance by a neural network that
can be trained with any of the standard supervised learning techniques. A
concrete example will be considered below, see \eqn{eqn:NeuralControlForward10}.
The relation to the kernel
notation is obtained by observing that
\begin{equation}
p\left(  s_t|s_{t-1}\right)  =\int\delta\left(  s_t-\psi\left(
s_{t-1}\right)  -\xi_t\right)  p_{\Xi}\left(  \xi_t\right) \dint\xi_t=p_{\Xi}\left(
s_t-\psi\left(  s_{t-1}\right)  \right)  \label{eqn:estimat:PI10}%
\end{equation}
where $\delta\left(  x\right)  $ is the Dirac delta distribution and $p_{\Xi
}\left(  \xi\right)  $ is the probability density of the random variable $\Xi$ (prediction error)
which may depend on the state $s$ itself (multiplicative noise).

The case of linear systems, where $\psi\left(  s\right)  =Ls+b$ with a
constant matrix $L$, has been treated in
\citet{AyDerBernigauProkopenko2012} revealing many interesting properties of
the PI. How can we translate the findings of the linear systems to the case of
nonlinear systems? As it turns out, the nonlinearities introduce many
difficulties into the evaluation of the PI as it becomes clear already in a
one-dimensional bistable system as treated in \citet{ay08:predinf_explore_behavior}.
Higher dimensional systems bring even more of such difficulties
 so that we propose to consider the information quantity on a new basis.
The idea is to study the TiPI of the error propagation dynamics in the
stochastic dynamical system instead of the process $S_{t}$ itself.

\subsubsection{Error propagation dynamics}
Let us introduce a new variable describing the deviation of the actual dynamics, \eqn{eqn:estimat:PI:psi},
 from the deterministic prediction in a certain time window.
We define for a time window starting at time $t-\tau$
\begin{equation}
\delta s_{t^{\prime}}=s_{t^{\prime}}-\psi^{t^{\prime}-\left(t-\tau\right)}\left(s_{t-\tau}\right)
\label{eqn:errorforward22}%
\end{equation}
for any time $t^{\prime}$ with $t-\tau\leq t^{\prime}\leq t$ and $\psi^{\left(0\right)}\left(s\right)=s$.
As to notation, $\delta s$ denotes a single variable not to be confused
 with the Dirac function.
Intuitively $\delta s_{t^{\prime}}$ captures
 how the prediction errors occurred since the start of the time window are propagated up to time $t'$.
\begin{figure}
\centering
\includegraphics[width=.7\textwidth]{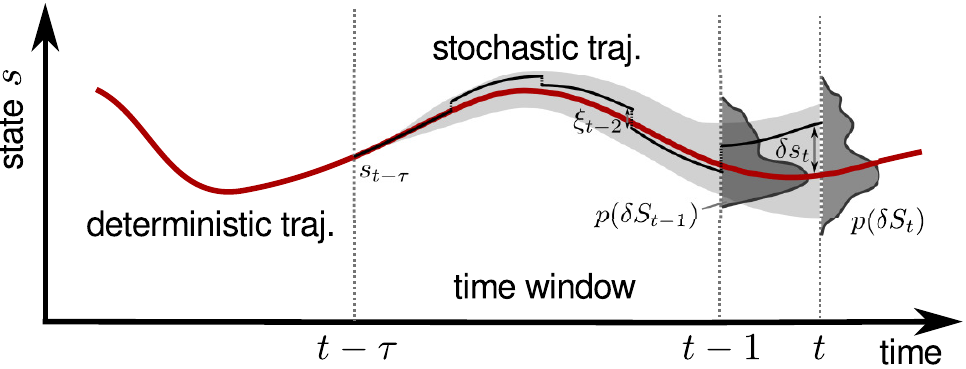}
\caption{{\bf The time window and the error propagation dynamics
  used for calculating the TiPI, \eqn{eqn:Idelta_gleich_IS}.}
In principle, the process is considered many times with always the same starting
value but different realizations of the noise $\xi$. Note that, when using the
one-shot gradients, only one realization is needed, see \sec{sec:LearningRules}.
}%
\label{fig:timewindow-errorprop}%
\end{figure}
\Fig{fig:timewindow-errorprop} illustrates the transformed state and the relevant
 distributions of the belonging process $\delta S_t$.
Interestingly the TiPI on the process $\delta S$ is equivalent to the one on the original
 process $S$, see \eqn{eqn:Idelta_gleich_IS}.
The dynamics of the $\delta s$ can be approximated by linearization as
\begin{equation}
\delta s_{t^{\prime}}=L\left(  s_{t^{\prime}-1}\right)  \delta s_{t^{\prime
}-1}+\xi_{t^{\prime}} + O(\|\xi_t\|^2) \label{eqn:errorforward:dyn}%
\end{equation}
using the Jacobian
\begin{equation}
  L_{ij}(s) = \frac{\partial \psi_i(s)}{\partial s_j}\label{eqn:L}
\end{equation}

Assuming the prediction errors (noise) $\xi$ to be both small and Gaussian we obtain
 an explicit expression for the TiPI on $\delta S$
\begin{equation}
\Itau\left(  \delta S_{t}:\delta S_{t-1}\right)  =\frac{1}{2}\ln\left\vert
\Sigma_{t}\right\vert -\frac{1}{2}\ln\left\vert D_{t}\right\vert
\label{eqn:PI:explicit10}%
\end{equation}
where $\Sigma=\left\langle \delta s \delta s^\T \right\rangle$ is the covariance matrix of $\delta S$
 and $D=\left\langle \xi \xi^\T \right\rangle$ is the covariance matrix of the noise.
The derivation and further details are in the Appendix \secA{sec:error:propagation}.
The results for linear systems in \citet{AyDerBernigauProkopenko2012}
 can be obtained from the general case considered here by $\tau \rightarrow\infty$.

When looking at \eqn{eqn:PI:explicit10} one sees that the entropies
are expressed in terms of covariance matrices. This is exact in the case of
Gaussian distributions. In the general case this may be considered as an
approximation to the true TiPI. Alternatively, we can also consider
\eqn{eqn:PI:explicit10} as the definition of a new objective function for any
process if we agree to measure variability not in terms of entropies but more
directly in terms of the covariance matrices.

\subsection{The exploration dynamics}\label{sec:LearningRules}

Our aim is the derivation of an algorithm driving the behavior of the agent toward
increasing TiPI. Let us assume that the function $\psi:\Real^{n}\rightarrow
\Real^{n}$ depends on a set of parameters $\theta$ so that we may write the
dynamics as
\begin{equation}
s_{t}=\psi\left(  s_{t-1},\theta_t\right)  +\xi_{t+1}
\label{eqn:LearningRulesIntro10}%
\end{equation}
For instance, if $\psi$ is a neural network as introduced further below, the
parameter set $\theta$ comprises just the synaptic weights and threshold values of the neurons.

\subsubsection{Gradient ascending the TiPI}

Based on the TiPI, \eqn{eqn:PI:explicit10}, a rule for the parameter dynamics is given by the gradient
step to be executed at each time $t$
\begin{equation}
\Delta\theta_t=\varepsilon\frac{\partial \Itau}{\partial\theta} =
\varepsilon\frac{\partial}{\partial\theta}\ln\left\vert \Sigma_t\right\vert
\label{eqn:LearningRulesIntro10a}%
\end{equation}
where $\varepsilon$ is the update rate and $\theta_{t+1} = \theta_t + \Delta \theta_t$.
The term $\ln|D|$ from  \eqn{eqn:PI:explicit10} has been omitted
 assuming that $\xi$ is essentially noise which is not depending on the parameters of the controller.
This is justifiable in the case of parsimonious
control as realized by the low-complexity controller networks.
These generate typically well predictable (low noise) behaviors as shown in the applications studied below.


In order to get more explicit expressions, let us consider the case of very short time windows. With $\tau=1$ there is
no learning signal since $\Sigma=D$ meaning that $\Itau=0$.
So, $\tau=2$ is the most simple nontrivial case.
The parameter dynamics is given by
\begin{equation}
\Delta\theta_t=\varepsilon\delta u_{t}^{\T}\frac{\partial L\left(  t-1\right)
}{\partial\theta}\delta s_{t-1} \label{eqn:LearningRulesSpecial24a}%
\end{equation}
where $\delta s$ and the auxiliary vector $\delta u$ are given as%
\begin{align}
\delta s_{t-1}  &  =s_{t-1}-\psi\left(  s_{t-2}\right)
\label{eqn:LearningRulesSpecial24aa}\\
\text{ }\delta s_{t}  &  =s_{t}-\psi\left(  \psi\left(  s_{t-2}\right)
\right)  \text{ }\label{eqn:LearningRulesSpecial24ab}\\
\delta u_{t}&=\Sigma_{t}^{-1}\delta s_{t}\komma
\label{eqn:LearningRulesSpecial24b}\\%
\Sigma_{t}  &  =\left\langle \delta s_{t}\delta s_{t}^{\T}\right\rangle
\label{eqn:LearningRulesSpecial24ac}%
\end{align}
stipulating the noise is different from zero (though possibly infinitesimal) and employing the self-averaging
 property of a stochastic gradient, see below.
The general parameter dynamics for arbitrary $\tau$ is derived in Appendix \secA{sec:LearningRules:General}.
However, in the applications described below, already the simple parameter dynamics with $\tau=2$ will be seen
to create most complex behaviors of the considered physical robots.

In a nutshell, \eqn{eqn:LearningRulesIntro10a} reveals already the main effect of TiPI
maximization: increasing $|\Sigma|$ means increasing the norm of $\delta s$
 (in the $\Sigma$-metric see \eqn{eqn:LearningRulesIntro30}).
This is achieved by increasing the
amplification of small fluctuations in the sensorimotor dynamics which is
equivalent to increasing the instability of the system dynamics, see also the
more elaborate discussion in~\citet{DerGuettlerAy08:predinf}.

\subsubsection{Learning vs.\ exploration dynamics}

Usually, updating the parameters of a system according to a given objective is
called learning. In that sense, the gradient ascent on the TiPI defines a
learning dynamics. However, we would like to avoid this notion here, since
actually nothing is learnt. Instead by the interplay between the system and
the parameter dynamics, the combined system never reaches a final behavior
corresponding to the goal of a learning process. Therefore we prefer the
notion \emph{exploration dynamics} for the dynamics in the parameter space that is
driven by the TiPI maximization.

\subsubsection{One-shot gradients}\label{sec:one-shot}

The formulas for the gradient (\eqns{eqn:LearningRulesSpecial24a}{eqn:LearningRulesIntro30}) were obtained by
tacitly invoking the self-averaging properties of the gradient, \ie by simply
replacing $\left\langle \delta s\delta s^{\T}\right\rangle $ with $\delta
s\delta s^{\T}$ in \eqn{eqn:LearningRulesIntro20a}. This still needs a little
discussion. Actually, the self-averaging is exactly valid only in the limit of
sufficiently small $\varepsilon$, with $\varepsilon$ eventually being driven
to zero in a convenient way. However, our scenario is different. What we are
aiming at is the derivation of an intrinsic mechanism for the self-determined and
self-directed exploration using the TiPI and related objectives. The essential
point is that self-exploration is driven by a
deterministic function of the states (sensor values) of the system itself.

\Eqn{eqn:LearningRulesSpecial24a} obtained from the gradient of the TiPI
fulfills these aims very well---any change of the system parameters and hence
of the behavior is given in terms of the predecessor states in the short time
window. With finite (and often quite large) $\varepsilon$
\eqnr{eqn:LearningRulesSpecial24a}{eqn:LearningRulesSpecial24ac} are just a
rough approximation of the original TiPI but, in view of our goal, the one-shot
nature of the gradient is favorable as it supports the explorative nature of
the exploration dynamics generating interesting synergy effects.

\subsubsection{Synergy of system and exploration dynamics}\label{sec:Synergy}

A further central aspect of our approach is the interplay between the system
and the parameter dynamics driven by the TiPI maximization process. In specific
cases, the latter may show convergence as in conventional approaches based on
stationary states. An example is given by the one-parameter system studied in
\citet{ay08:predinf_explore_behavior} realizing convergence to the so called
effective bifurcation point. However, with a richer parametrization and/or
more complex systems, instead of convergence, the combined system (state +
parameter dynamics) never comes to a steady state due to the intensive
interplay between the two dynamical components if $\varepsilon$ is kept finite.
An example will be given in the Results section. 

Typically, the TiPI landscape permanently changes its shape due to the fact that
increasing the TiPI means in general a destabilization of the system dynamics.
If the latter is in an attractor, increasing the TiPI destabilizes the attractor
until it may disappear altogether with a complete restructuring of the TiPI
landscape. This is but one of the possible scenarios where the exploration
dynamics engages into an intensive and persistent interplay with the system
dynamics. This interplay leads to many synergistic effects between system and
exploration dynamics and makes the actual flavor of the method.

\subsubsection{Self-directed search}

The common approach to solve the exploration--exploitation dilemma in learning
problems is to use some randomization of actions in order to get the necessary
exploration and then decrease the randomness to exploit the skills acquired so far.
This is prone to the curse of dimensionality if the systems are gaining some
complexity.
Randomness can also be introduced by using a deterministic policy with a random component in the parameters, as quite successfully applied to evolution strategies and reinforcement learning~\citep{DBLP:journals/nn/SehnkeORGPS10,hansen2001ecj}.

Our approach is also to use deterministic policies
 (given by the function $K$) but aims at making
exploration part of the policy. So, instead of relegating exploration to the
obscure activities of a random number generator, variation of actions should
be generated by the responses of the system itself. This replaces randomness
with spontaneity and is hoped (and will be demonstrated) to restrict the
search space automatically to the physically relevant dimensions defined by
the embodiment of the system.

Formally, we call a search self-directed
if there exists a function $\alpha$ so that the change in the parameters
\begin{equation}
\Delta \theta_{t}=\alpha\left(  s_{t},\ldots,s_{t-\tau},\theta_{t}\right)
\label{eqn:Method:questions20}%
\end{equation}
is given as a deterministic function of the states in a certain time window (of length  $\tau$)
 and the parameter set $\theta$ itself. In this paper, $\alpha$ is given by the gradient of the
 predictive information in the one-shot formulation, see \sec{sec:one-shot}.

In more general terms, we believe that randomization of actions makes the
agent heteronomous, its fate being determined by an obscure (to him)
 procedure (the pseudo-random number generator) alien to the
 nature of its dynamics. The agent is autonomous in the 'genuine' sense only if it varies its actions exclusively by its own internal laws~\citep{RohdeS08}. In our approach,
according to \eqn{eqn:Method:questions20}, exploration is driven entirely by
the dynamics of the system itself so that exploration is coupled in an
intimate way to the pattern of behavior the robot is currently in.
The danger might be that in this way the exploration is restricted too much.
As our experiments show, this is not so for active motion patterns in high dimensional
systems. This fact can be attributed to the destabilization effect
incurred by the TiPI maximization, see above and~\citep{DerGuettlerAy08:predinf}. For stabilizing behaviors, however, the exploration may be too restrictive.

\subsection{The sensorimotor loop}

Let us now specify the above expressions to the case of a sensorimotor loop,
in particular a neurally controlled robotic system. The dynamical systems
formulation is obtained now by writing our predictor for the next sensor
values as a function of both the sensors and the actions so that
\begin{equation}
s_{t}=\phi\left(  s_{t-1},a_{t-1}\right)  +\xi_{t}
\label{eqn:InfTheory:SMdynPhi}%
\end{equation}
where $\phi$ represents the so-called forward model and $\xi_{t}$ is the
prediction error as before. As the next step, we consider the controller also
as a deterministic function $K:\Real^{n}\rightarrow \Real^{m}$ generating actions
(motor values) $a_{t}\in \Real^{m}$ as a function of the sensor values $s_{t}\in
\Real^{n}$ so that
\begin{equation}
a_t=K\left(  s_t\right)  \period \label{eqn:InfTheory:SMdynK}%
\end{equation}
In the applications, $K$ will be realized as a (feed-forward) neural network.
Using \eqn{eqn:InfTheory:SMdynK} in \eqn{eqn:InfTheory:SMdynPhi} we obtain the
map $\psi$ modeling our sensor process as
\begin{equation}
\psi\left(  s_{t-1}\right)  =\phi\left(  s_{t-1},K\left(  s_{t-1}\right)
\right) \label{eqn:SML:psi} \period
\end{equation}
In~\citet{AyDerBernigauProkopenko2012} a standard linear control system was
studied where
$K\left(  s\right)  =Cs\text{, \ }\phi\left(  s,a\right)  =Ts+Va$ and
$\psi\left(  s\right)  =\left(  T+VC\right)  s$.
This paper will consider a nonlinear generalization of that case in specific
robotic applications.

\subsubsection{Exploration dynamics for neural control systems}

In the present setting, we assume that both the controller $K$ and the forward
model $\phi$ of our robot are realized by neural networks, the controller
being given by a single-layer neural network as
\begin{equation}
K\left(  s\right)  =g\left(  Cs+h\right)  \label{eqn:NeuralControl10}%
\end{equation}
the set of parameters $\theta$ now given by $C$ and $h$.
In the concrete applications to be given below, we specifically use $g_i\left(
z\right)  =\tanh\left(  z_i\right)  $ (to be understood as a vector function so
that $g:\Real^{n}\rightarrow \Real^{n}$).

Moreover, the forward model $\phi$ is given by a layer of linear neurons, so
that
\begin{equation}
\phi\left(  s,a\right)  =Va+Ts+b\period \label{eqn:NeuralControl:psi}%
\end{equation}
The matrices $V$, $T$ and the vector $b$
represent the parametrization of the forward model that is adapted on-line
by a supervised gradient procedure to minimize the prediction error $\xi^\T\xi$ as
\begin{equation}
\Delta V=\eta_\phi\xi a^{\T}\text{,\ \ }\Delta T=\eta_\phi\xi s^{\T}\text{,\ \ }\Delta
b=\eta_\phi\xi \period\label{eqn:NeuralControlForward10}%
\end{equation}
In the applications, the learning rate $\eta_\phi$
is large such that the low complexity of the model is compensated by a very
fast adaptation process.

In contrast to the forward model parameters,
 the controller parameters are to be adapted to maximize the TiPI.
For that the map $\psi$ \eqnp{eqn:SML:psi} is required which
becomes $\psi\left(  s\right)  =Vg\left(  Cs+h\right)  +Ts+b$
with Jacobian matrix
\begin{equation}
L=VG^{\prime}\left(  z\right)  C+T \label{eqn:NeuralControlForward10L}%
\end{equation}
where $z=Cs+h$ is the postsynaptic potential and
\begin{equation}
G'(z)=\textrm{diag}[g'(z_1),\dots, g'(z_m)]\label{eqn:NeuralControl:G}
\end{equation}
is the diagonal matrix of the derivatives of the activation functions for each control neuron.

In the applications given below, we are using the short-time window, with the
general exploration dynamics given by \eqn{eqn:LearningRulesSpecial24a}.
The explicit exploration dynamics for this neural setting with $g(z)=\tanh(z)$
 are given as
\begin{align}
\frac{1}{\varepsilon}\Delta C_{ij}  &  =\delta\mu_{i}\delta s_{j}-\gamma
_{i}a_{i}s_{j}\komma\label{eqn:NeuralControl:LRa}\\
\frac{1}{\varepsilon}\Delta h_{i}  &  =-\gamma_{i}a_{i}
\label{eqn:NeuralControl:LRb}%
\end{align}
where all variables are time dependent and are at time $t$, except $\delta s$ which is at time $t-1$.
The vector $\delta\mu\in \Real^{m}$ is defined as
\begin{equation}
\delta\mu_t=G^{\prime}V^{\T}\delta u_{t}=G^{\prime}V^{\T}\Sigma^{-1}\delta s_{t}
\label{eqn:NeuralControl:mue}%
\end{equation}
(see \eqn{eqn:LearningRulesSpecial24b}), and the channel specific learning
rates $\gamma_{i}$ are
\begin{equation}
\gamma_{i}=2\left(  C\delta s_{t-1}\right)  _{i}\delta\mu_{i}\period
\label{eqn:NeuralControl:gamma}%
\end{equation}
The derivation and generalization to aribrary activation functions are provided in the
Appendix \secA{SecAppendix:NeuralLearn}.
The update rules for $\tau>2$ are given by a sum of such terms, with appropriate
redefinitions of the vector $\delta\mu$, see \eqn{eqn:AppLearningRuleGeneral10final}.

\subsubsection{The Hebbian nature of the update rules}

In order to interpret these rules in more neural terms, we at first note that
the last term in \eqn{eqn:NeuralControl:LRa} is of an anti-Hebbian structure.
In fact, it is given by the product of the output value $a_{i}$ of neuron
$i$ times the input $s_{j}$ into the $j$-th synapse of that neuron, the
$\gamma_{i}$ (which are positive, as a rule) being interpreted as a neuron
specific learning rate. Moreover, we may also consider the term $\delta\mu
_{i}\delta s_{j}$ as a kind of Hebbian since it is again given by a product of
values that are present at the ports of the synapse $j$ of neuron $i$. The
factor $\delta s_{j}$ can be considered as a signal directly feeding into the
input side of the synapse $C_{ij}$. Moreover, $\delta\mu$ given as $\delta
\mu=G^{\prime}V^{\T}\delta u$ is obtained by using $\delta u$ as the vector of
output errors in the $\psi$ network and propagating this error back to the
layer of the motor neurons by means of the standard backpropagation algorithm.

These results make the generalization to more complicated, multi-layer
networks straightforward. However already the simple setting
 produces an overwhelming behavioral variety, see the case studies below.

More intuitively the Hebbian term acts as a self-amplification and
 increases the Lyapunov exponents.
In the linear case~\citep{AyDerBernigauProkopenko2012} this leads eventually to the divergence
 of the dynamics such that the PI does not exist any longer. With the
nonlinearities, the latter effect is avoided, but the system is driven into the
saturation region of the motor neurons. However, the second term in \eqn{eqn:NeuralControl:LRa},
by its anti-Hebbian nature, is seen to counteract this tendency. The net effect of
both terms is to drive the motor neurons towards a working regime where the
reaction of the motors to the changes in sensor values is maximal. This is
understandable, given that maximum entropy in the sensor values requires a
high sensorial variety that can be achieved by that strategy.

\section{Results}

We apply our theory to three case studies to illuminate the main features.
First a hysteresis systems is considered to exemplify the consequences of
 nonstationarity and the resulting interplay
 between the exploration dynamics and the system dynamics in a nutshell.
In \sec{sec:spontancoop}
 a physical system of many degrees of freedom is controlled by independent
 controllers that spontaneously cooperate.
Finally in \sec{sec:Humanoid}
 we apply the method to a jointly controlled humanoid robot in various situations
 to illustrate the exploration process in a high-dimensional embodied system.

\subsection{Hysteresis systems}

Nonstationary processes are the main target of our theory, made
accessible by the special windowing and averaging technique presented in this
 paper for the first time.
In order to work out the consequences,
let us consider an idealized situation where the above derivations, in particular
\eqnr{eqn:LearningRulesSpecial24a}{eqn:LearningRulesSpecial24ac}, are the
exact update rules for increasing the TiPI.

Let us consider a single neuron in an idealized sensorimotor loop,
 where the sensor values are
$s_{t}=a_{t-1}+\xi_{t}$ (the white Gaussian noise $\xi$ is added explicitly).
This case corresponds to the dynamical system
\begin{equation}
s_{t}=\tanh\left(  Cs_{t-1}+h\right)  +\xi_{t}
\label{eqn:hysteresissystems10}%
\end{equation}
where now $s_{t}\in \Real^{1}$. The system was studied earlier
\citep{DerGuettlerAy08:predinf} in the special case of $h=0$ and it was shown
that the maximization of the PI self-regulates the system parameter $C$
towards a slightly supercritical value $(1 < C \ll 2)$. There, the system is at the so
called effective bifurcation point where it is bistable but still sensitive to
the noise.

\begin{figure}
\begin{center}
\includegraphics[scale=0.75]{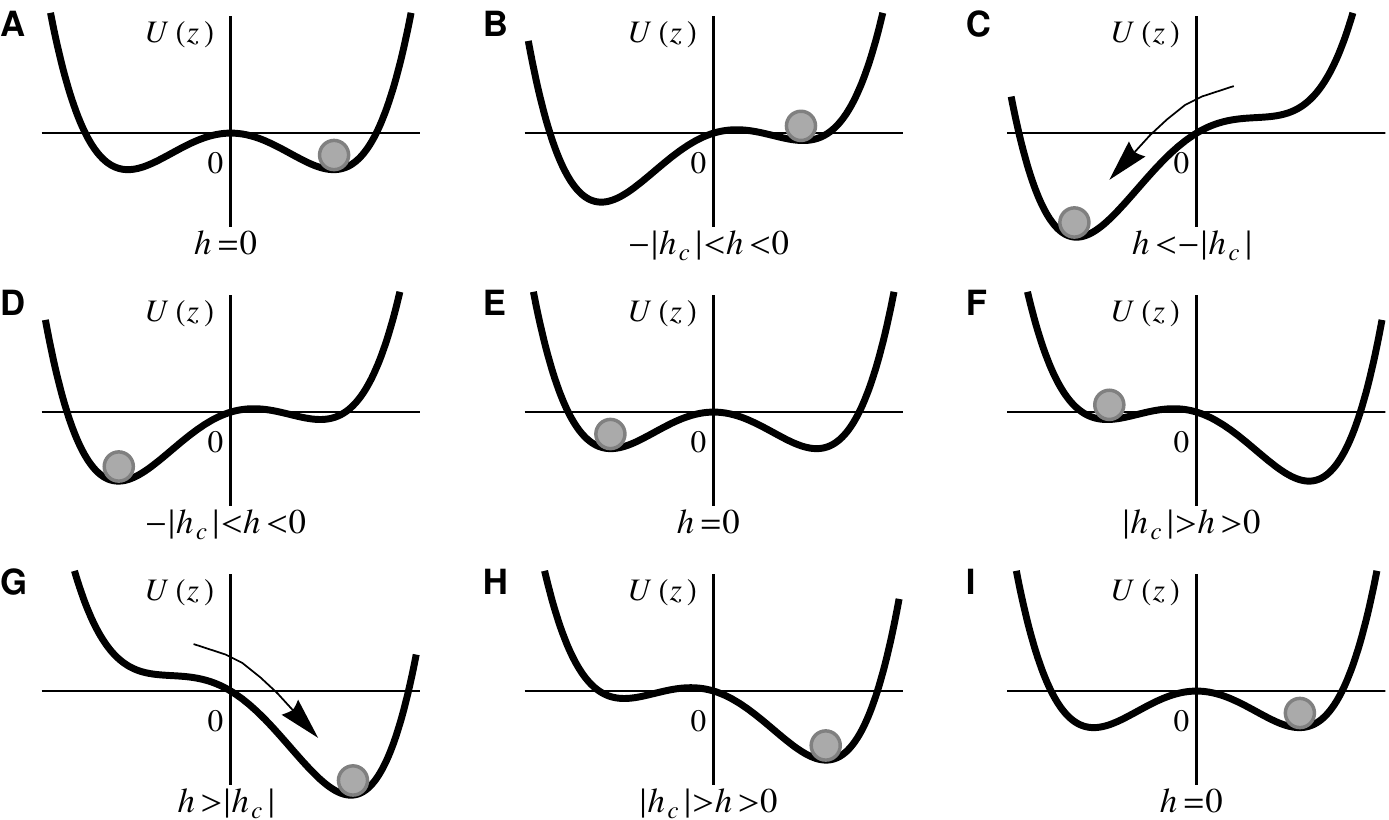}
\end{center}
\caption{{\bf The hysteresis cycle in the gradient picture.} The diagrams show the
stages of one hysteresis cycle starting from $h=0$ (\infig{A}) with the state
at $z>0$ as represented by the sphere. Decreasing $h$ creates the asymmetric
situation (\infig{B}). If $h=-h_{c}$ the saddle-node bifurcation happens,
\ie both the maximum at z=0 and the right minimum disappear so that the system
shifts to the left minimum of the potential (\infig{C}). Increasing $h$ until
$h=0$ brings us back to the initial situation with the state shifted to the
other well see (\infig{D,E}). The diagrams (\infig{F}) and (\infig{G}) depict
the switching from the minimum at $z<0$ to the minimum at $z>0$ by increasing
$h$. By decreasing $h$ until $h=0$ the hysteresis cycle is finished, see
(\infig{H,I}). }%
\label{fig:SensMotLoop:hysterese_potential}%
\end{figure}

Let us start with keeping $C$ fixed at some supercritical value (\eg $C=1.1$) and concentrating on
the behavior of the bistable system as a function of the threshold value $h$.
The interesting point is that the system shows hysteresis. This can be
demonstrated best by rewriting the dynamics in state space as a gradient
descent. Let us introduce the postsynaptic potential $z_{t}=Cs_{t}+h$ and
rewrite \eqn{eqn:hysteresissystems10} in terms of $z_{t}$ as
\begin{equation}
\Delta z_{t}=-\frac{\partial}{\partial z_{t}}U\left(  z_{t}\right)  +\xi_{t+1}%
\end{equation}
where $\Delta z_{t}=z_{t+1}-z_{t}$ and the potential is $U\left(  z\right)
=-C\ln\cosh z - h z + \frac{z^{2}}{2}$ (using $\frac{\partial}{\partial z}\ln\cosh
z=\tanh z$).
In that picture, the hysteresis properties of the system are most easily
demonstrated by \fig{fig:SensMotLoop:hysterese_potential}.
This phenomenon can be related directly to the destabilization
effect of the exploration dynamics. In the potential picture, stability is
increasing with the well depth. Hence, the exploration dynamics, aiming at the
destabilization of the system, is decreasing the depth of the well more and
more until the well disappears altogether, see
\fig{fig:SensMotLoop:hysterese_potential}, and the state switches to the other well
where the procedure restarts.

\subsubsection{Deterministic self-induced hysteresis oscillation}

Now we show that in the one-dimensional case the parameter dynamics is
 independent of white noise. This implies we can in the state dynamics
 make the limit of vanishing noise strength and obtain a fully deterministic system.
Again we only consider the two-step window ($\tau=2$).
Using $\delta s_{t}=\xi_{t}+L\delta s_{t-1}=\xi_{t}+L\xi_{t-1}$ \eqnp{eqn:errorforward:dyn} we find that the TiPI,
 according to \eqn{eqn:PI explicittau2}
\[
\Itau=\frac{1}{2}\ln\left(  1+L^{2}\right)  \komma
\]
is independent of the noise.
Analogously to \eqnr{eqn:NeuralControl:LRa}{eqn:NeuralControl:gamma}
 we obtain the update rules for $C$ and $h$ as
 the gradient ascent on $\Itau$ and thus the full state-parameter dynamics (with $|\xi| \to 0$) is given by
\begin{align}
s_{t}  &  =g\left(  C_{t-1}s_{t-1}+h_{t-1}\right)  \komma\label{eqn:one-dim-learn:state}
\\
C_{t}  &  =C_{t-1}+\gamma\left(  1/(2C_{t-1}) - s_{t-1}a_{t-1}\right)\label{eqn:one-dim-learn:C}\\
h_{t}  &  =h_{t-1}-\gamma s_{t}\komma \label{eqn:one-dim-learn:h}
\end{align}
with $\gamma = 2 L^{2} /(1+L^{2})$.

\begin{figure}
\centering
\begin{tabular}{cc}
\infig{A}\ \ $C=1.2$ & \infig{B}\ \ full dynamics\\
\includegraphics[width=0.45\linewidth]{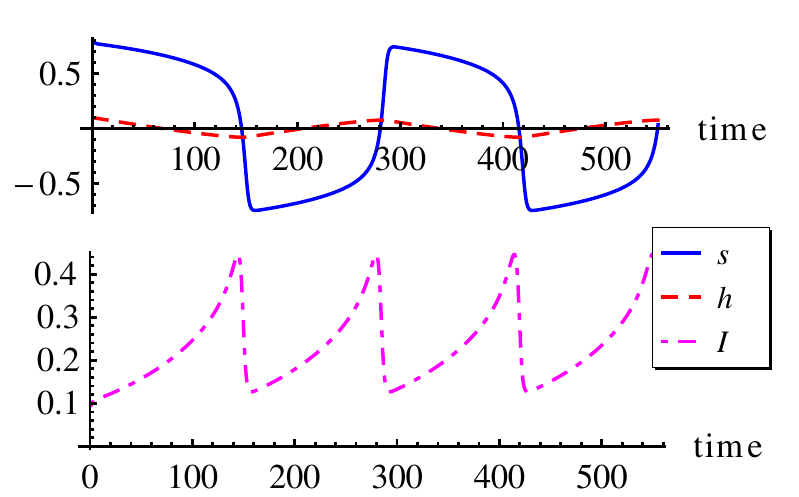}&
\includegraphics[width=0.45\linewidth]{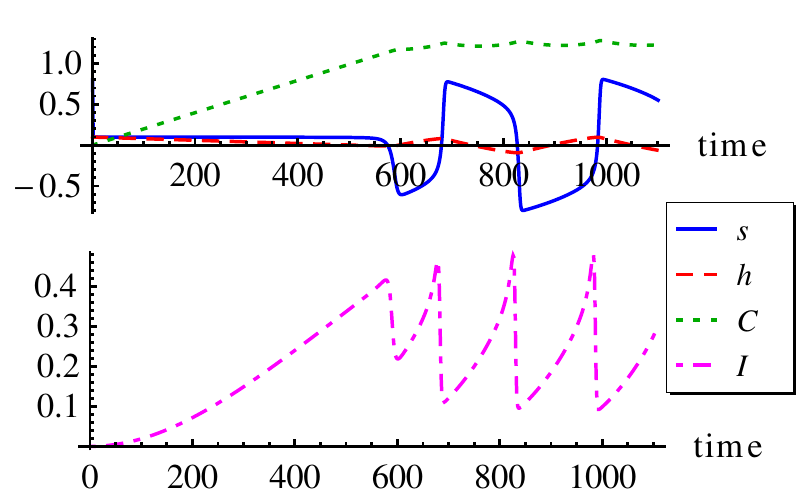}
\end{tabular}
\caption{{\bf State and parameter dynamics in the one-dimensional system.}
(\infig{A}) Only $h$ dynamics (fixed $C=1.2$); the bias $h$ oscillates around
zero and causes the state $s$ to jump between the positive and negative fixed
points. The TiPI is seen to increase steadily until it eventually drops back
when the state is jumping. (\infig{B}) With full dynamics ($C,h$). $C$
increases until it oscillates around its average at $C\approx1.2$ where the
hysteresis cycle starts. Parameters: $h_{0}=0.1$, $s_{0}=0.8$, $C_{0}=0$ in (\infig{B}),
$\varepsilon=0.002$ }%
\label{fig:HKEx:1d:dynamics}%
\end{figure}

Apart from the definition of $\gamma$ (that just modulates the
speed of the parameter dynamics), the extended dynamical system agrees in the
one-dimensional case with that derived from the principle of homeokinesis,
 discussed in detail in \citet{DerMartius11}.
Let us therefore only briefly sketch the most salient
features of the dynamics. Keeping $C$ fixed at some supercritical value,
 as above the most important point is that, instead of
converging towards a state of maximum TiPI, the $h$ dynamics drives the neuron
through its hysteresis cycle as shown in \fig{fig:SensMotLoop:hysterese_potential},
 which we call a self-induced hysteresis oscillation, see \fig{fig:HKEx:1d:dynamics}(A).

For the full dynamics (with \eqn{eqn:one-dim-learn:C}) the
results are given in \fig{fig:HKEx:1d:dynamics}(B) showing that the
feedback strength $C$ in the loop converges indeed toward the regime with the
hysteresis oscillation. This demonstrates that the latter is not an artifact
present only under the specific parametrization. In fact, we encounter this
phenomenon in many applications with complex high-dimensional robotic systems,
see the experiments with the \Armband{} below and many examples
treated in~\citet{DerMartius11}.

Interestingly this behavior is not restricted to simple hysteresis systems
 but is of more general relevance. For instance, in two-dimensional systems a second
order hysteresis was observed, corresponding to a sweep through the frequency
space of the self-induced oscillations~\citep{DerMartius11}. It would be
interesting to relate this fast synaptic dynamics to the spike-timing-dependent
plasticity~\citep{Markram97:STDP} or other plasticity rules~\citep{turrigiano98}
 found in the brain.

\subsubsection{About time windows}

\begin{figure}
\centering
\includegraphics[scale=1.1]{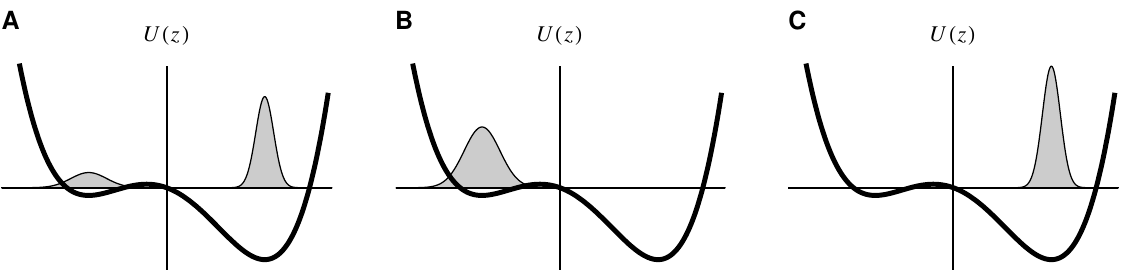}
\caption{{\bf The probability density distributions with different time windows of
the stochastic process in an asymmetric double well potential.} The mean first
passage time $T_{f}$ of switching between wells is one characteristic time
constant of the process~\citep{Risken89}, $T_{f}$ increasing exponentially with
the barrier height. If observing the process in a window of length $T \gg
T_{f}$, the distribution of \infig{(A)} will be observed. In that situation, the TiPI is
maximal if the wells are of equal depth ($h=0$). However, with windows of
length $T\ll T_{f}$, the system state will be predominantly in one of the wells
generating the distributions shown in \infig{(B)}, \infig{(C)}.
Gradient ascending the TiPI will
decrease the well depth as long as the probability mass is still concentrated
in that well. This is what drives the hysteresis cycle depicted in
\fig{fig:SensMotLoop:hysterese_potential}.}%
\label{fig:doublewell_distri}%
\end{figure}

Before giving the applications to embodied systems,
 let us have a few remarks on the special
 nature of the time windowing technique as compared to the common settings.
Let us consider again the bistable system with the bias $h$ as the only parameter
and with finite noise. \Fig{fig:doublewell_distri} depicts a typical situation
with $h\neq0$ so that the wells are of different depth. The figure depicts the
qualitative difference between the classical attitude of considering
information measures in very large time windows, large enough for the process
to reach total equilibrium, as compared to our nonstationarity approach where
the TiPI is estimated on the basis of a comparatively short window\footnote{Note
that the time to stay in a well is exponentially increasing with the depth of
the well and decreasing exponentially with the strength of the
noise~\citep{Risken89}. Mean first passage times can readily exceed physical
times (on the time scale of the behavior) by orders of magnitudes.}.

While in the former case convergence of the hysteresis parameter $h$ towards
the equilibrium condition $h=0$ is reached, there is no convergence in the
nonstationary case. Instead, one obtains a self-induced hysteresis
oscillation. This is generic for a large class of phenomena based on the
synergy effects between system and exploration dynamics, see \sec{sec:Synergy},
which open new horizons for the explorative capabilities of the agent. In the
context of homeokinesis, this phenomenon has already been investigated in many
applications, see~\citet{DerMartius11}. This paper provides a new, information theoretic basis and opens
new horizons for applications as the matrix inversions inherent to the
homeokinesis approach are avoided.

\subsection{Spontaneous cooperation with decentralized control}\label{sec:spontancoop}

Let us now give examples illustrating the specific properties of the present
approach. We start with an example of strongly decentralized control where
the TiPI driven parameter dynamics leads to the emergence of collective modes.
Earlier papers have already demonstrated this phenomenon for a chain of
passively coupled mobile robots~\citep{ay08:predinf_explore_behavior,DerGuettlerAy08:predinf,ZahediAyDer2010:HigherCoordination}. In the setting
of~\citet{ay08:predinf_explore_behavior,DerGuettlerAy08:predinf},
each wheel was
being controlled by a single neuron with a synapse of strength $C$ defining
the feedback strength in each of the sensorimotor loops. There was no bias. As
it turned out, the TiPI in the sensorimotor loop is maximal if the synaptic strength $C$
is at its critical value where the system is bistable but still reacts to the
external perturbations, \ie the system is at its so-called effective
bifurcation point~\citep{DerMartius11}. As compared to the present setting,
these results correspond to using a time window of infinite length,
stipulating the presence of a stationary state.

The situation is entirely different when using the short time window and large
update rates allowing for the synergy effects. In experiments with the robot
chain, we observe better cooperativity with the hysteresis oscillations and
better exploration capabilities. The reason can be seen in the fact that the
self-regulated bias oscillations help the chain to better get out of impasse
situations. We do not give details here, since we will study in the following
an example that demonstrates the synergy effects even more convincingly.

\subsubsection{The \Armband{}}

\begin{figure}
\begin{center}
\includegraphics[width=.6\textwidth]{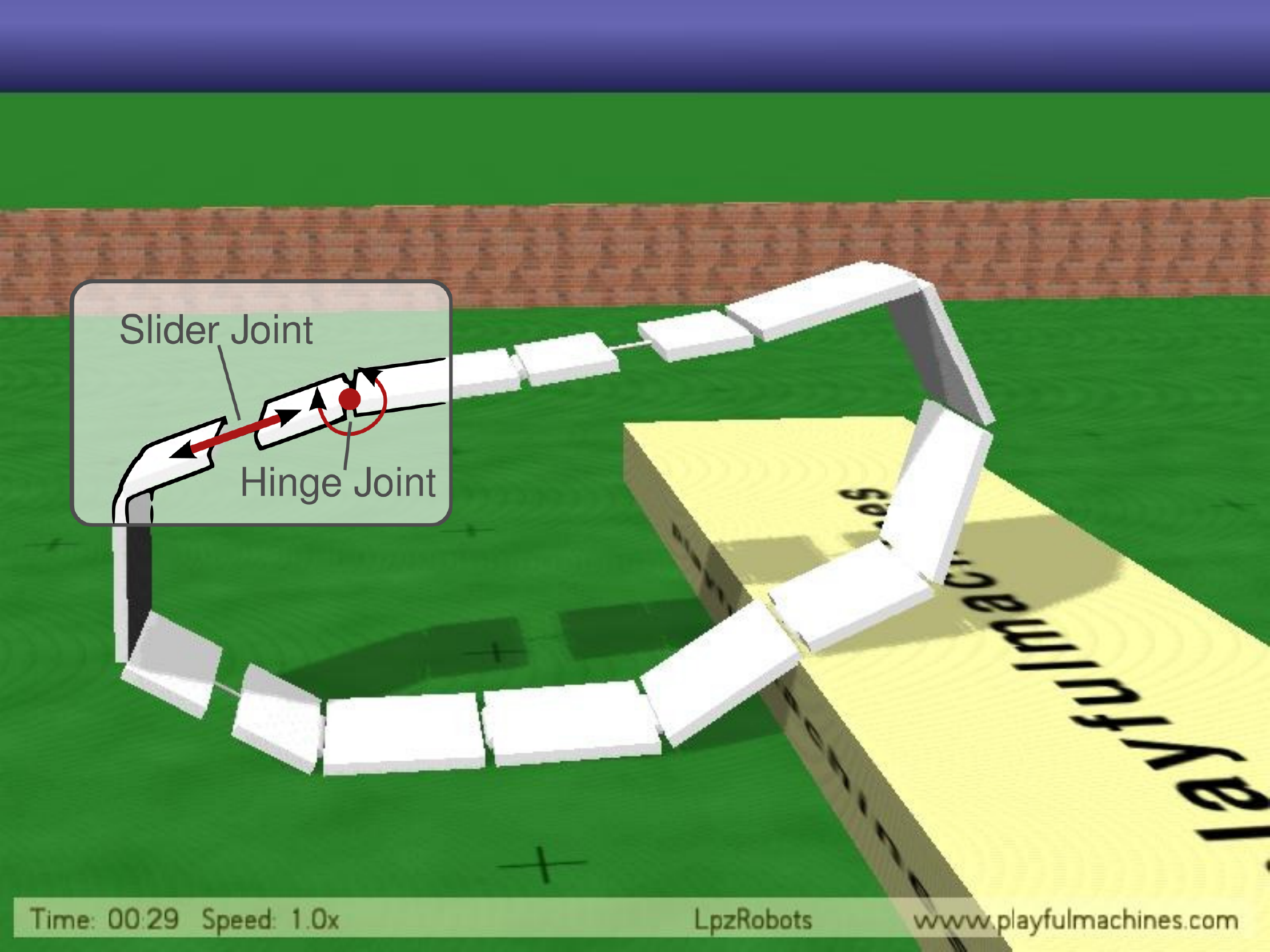}
\end{center}
\caption{{\bf The \Armband{}.} The robot has 12 hinge and 6 slider joints, each
actuated by a servo motor and equipped with a proprioceptive sensor measuring
the joint angle or slider length. The robot is strongly underactuated so that
it can not take on a wheel like form where locomotion were trivial.}%
\label{fig:Armband}%
\end{figure}

The \Armband{} considered here is a complicated physical object with $18$ degrees of freedom,
see \fig{fig:Armband}. The physics of the robot is
simulated realistically in the \lpzrobots{} simulator~\citep{lpzrobots10}.
The program source code for this and the next simulation is available from~\cite{videos:thispaper}.
Each joint is controlled by an individual controller, a
single neuron driven by TiPI maximization, as with the robot chain treated in
\citet{DerGuettlerAy08:predinf}. The controller receives the measured joint
angle or slider position and the output of the controller defines the target
joint angle or target slider position to be realized by the motor. The motors
are implemented as simulated servomotors in order to be as close to reality as
possible. Moreover, the forces are limited so that, due to the interaction
with obstacles or the entanglement of the system's different degrees of
freedom, the true joint angle may differ substantially from the target angle.
These deviations drive the interplay between system and
exploration dynamics.

In the experiments, we use the controller given by \eqn{eqn:NeuralControl10} and the update rules for the parameter dynamics as
given by \eqns{eqn:one-dim-learn:C}{eqn:one-dim-learn:h}.
The adaptive forward model is given by \eqn{eqn:NeuralControl:psi} with $T=0$ and the appropriate
 learning rules \eqn{eqn:NeuralControlForward10}.
In order to
demonstrate the constitutive role of the synergy effect, we started by
studying the system with fixed $C$ and $h=0$. In contrast to the chain of
mobile robots, with fixed parameters there is no parameter regime where the \Armband{} shows
substantial locomotion. This result suggests that, as compared to the chain of mobile robots,
the specific embodiment of the \Armband{} is more demanding
for the emergence of the collective effect.

In order to assess the effects appropriately, note that potential locomotion
 depends on the forces the motors are able to realize.
For instance, if the robot is strongly actuated, the command $a=0$ for each of the
motors drives each joint to its center position so that the shape of the
robot is nearly circular, locomotion readily taking place under the
influence of very weak external influences. In order to avoid such trivial
effects, we use an underactuated setting so that gravitational or
environmental forces are deforming the robot substantially, see
\fig{fig:Armband}.

\begin{figure}
\centering
\includegraphics[scale=0.8]{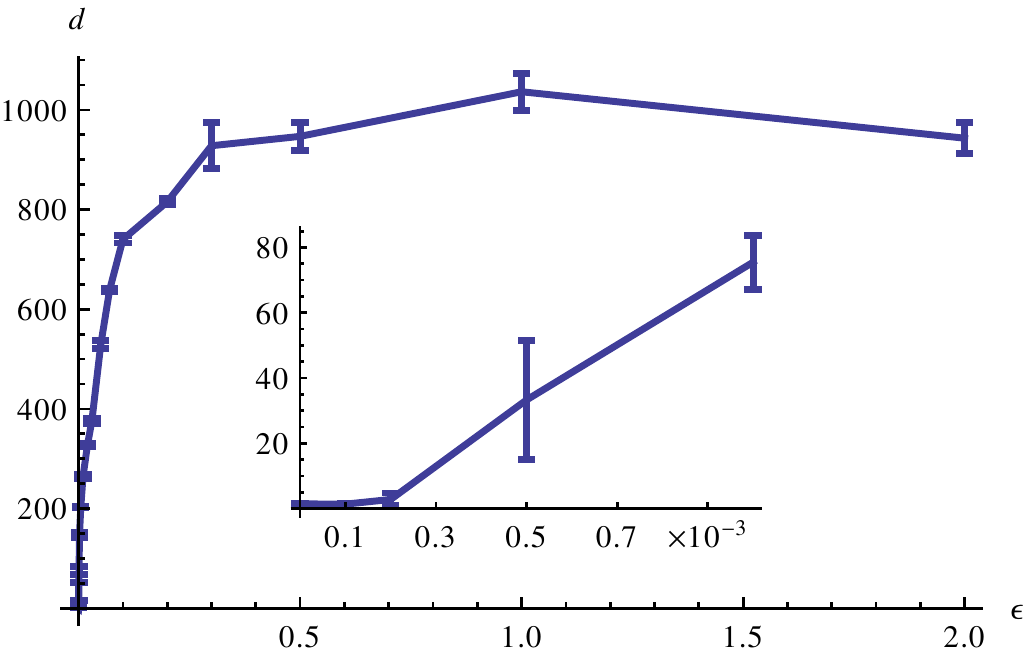}
\caption{{\bf Role of the fast synaptic dynamics: depending on the speed of the
synaptic dynamics defined by $\varepsilon$, the locomotion properties are
changing drastically.} Depicted is the distance traveled by the robot in
10\,min simulated time on an empty plane.
The inset gives a close up view for low $\varepsilon$,
demonstrating that the locomotion starts only if $\varepsilon$ exceeds a
certain threshold value. Shown is the mean and standard deviation of 10 runs
each.}%
\label{fig:armbandLocomotion}%
\end{figure}

\begin{figure}
\centering
\includegraphics[scale=0.8]{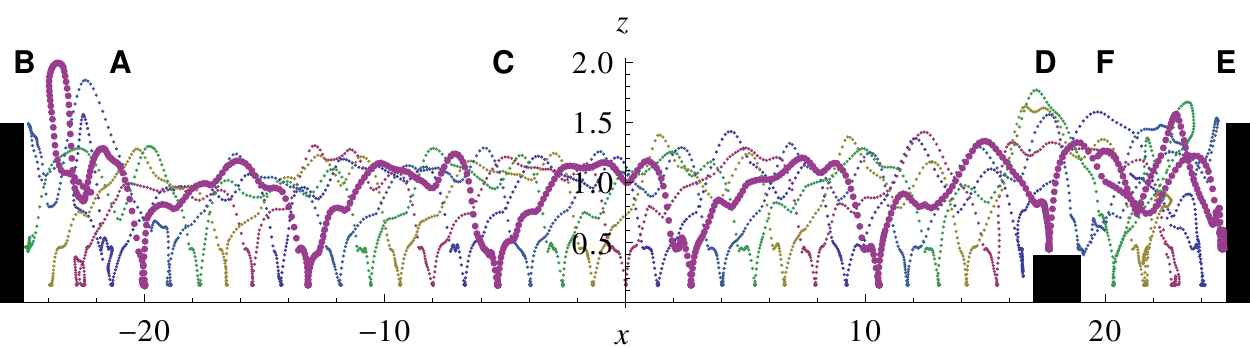}
\caption{{\bf Regular locomotion pattern and interaction with the environment.}
  Plotted are the center positions of the 6 rigid segments in space
  for an interval of 40\, sec. One line is highlighted for visibility.
  The trajectory starts while the robot is moving to the left (\infig{A}) and is hitting the wall (\infig{B}) (black box) and locomotes to the right (\infig{C}) showing a very regular pattern. Then it overcomes an obstacle (\infig{D}) and hits the wall (\infig{E}) and moves back (\infig{F}). The behavior is cyclic. Parameter: $\varepsilon=0.5$.
}%
\label{fig:armband:environment}%
\end{figure}

\begin{figure}
\centering
\includegraphics[width=1\linewidth]{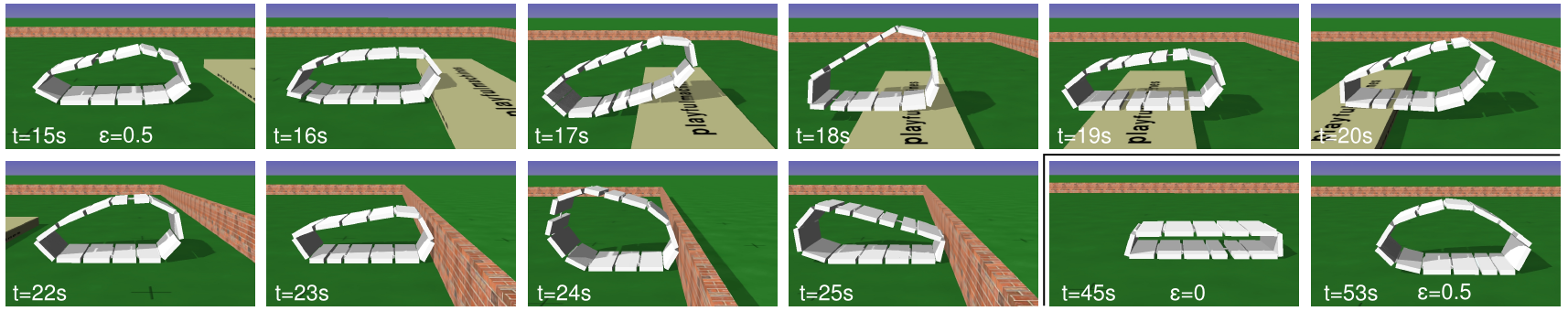}
\caption{{\bf \Armband{} robot surmounting an obstacle and inverting speed at a wall.}
Screen shots from the simulation for \fig{fig:armband:environment}. The order is row-wise from left to right.
The last two pictures show the situation after switching off the parameter dynamics
 $\varepsilon=0$ for a few seconds (the robots stops) and enabling it again (starts moving).
}%
\label{fig:armband:video}%
\end{figure}

The situation changes drastically if the $h$ dynamics is included. As
demonstrated by \fig{fig:armbandLocomotion}, substantial locomotion sets in
 only if $\varepsilon$ is large enough so that the exploration dynamics
is sufficiently fast for the synergy effect to unfold. Also, as the
experiments show, the effect is stable for a very wide range of $\varepsilon$
and under varying external conditions. It is also notable, that the \Armband{}
robot shows a definite reaction to external influences. For instance,
obstacles in its path are either surmounted or cause the robot to invert its
velocity, see \fig{fig:armband:environment}. The latter effect is observed in particular in the underactuated
regime defined above, so that the reflection is not the result of the elastic
collision but it is actively controlled by the involvement of the exploration dynamics.
The role of the latter is also demonstrated by the fact that
locomotion stops as soon as the update rate $\varepsilon$ is put to
zero, see \fig{fig:armband:video} and the corresponding video~\suppl{S1}.


The \Armband{} has also been investigated recently using artificial
evolution for the controller~\citep{Rempis_thesis}, demonstrating convincingly
the usefulness of the evolution strategy for obtaining recurrent neural
networks that make the \Armband{} roll into a given direction. There are several
differences to our approach, both conceptually and in the results. While in
the evolution strategy the fitness function was designed for the specific task
and many generations were necessary to get the performance, in our approach
the rolling modes are emerging right away by themselves. Moreover, the modes
are sensitive to the environment, for instance by inverting velocity upon
collisions with a wall, they are flexible (changing to a jumping behavior on
several occasions) and resilient under widely differing physical conditions.
Interestingly, these behaviors are achieved with an extremely simple neural
controller, the functionality of a recurrent network being substituted by the
fast synaptic dynamics.
%

\subsection{High dimensional case -- the \Humanoid{}}\label{sec:Humanoid}

Let us now study the properties of the exploration dynamics in a general (not
decentralized) control task. We consider a humanoid robot with $17$ degrees of
freedom. Each joint is driven by a simulated servo motor, the motor values
$a\in \Real^{17}$ sent by the controller are the target angles of the joints and
sensor values $s\in \Real^{17}$ are the true, observed angles. This is the only
knowledge the robot has about its physical state.

The aim of this experiment is to investigate in how far the robot develops
behaviors with high variability so that it explores its sensorimotor
contingencies. Given that there is no externally defined goal for the behavior
development, will the robot develop a high behavioral variety depending on its
physics and the environment it is dynamically embedded into?

\begin{figure}
\centering
\begin{tabular}
[c]{cccc}%
\infig{A} normal & \infig{B} bungee & \infig{C} high bar & \infig{D}
pit\\
\begin{minipage}[b]{2.2cm} \includegraphics[height=1.68cm]{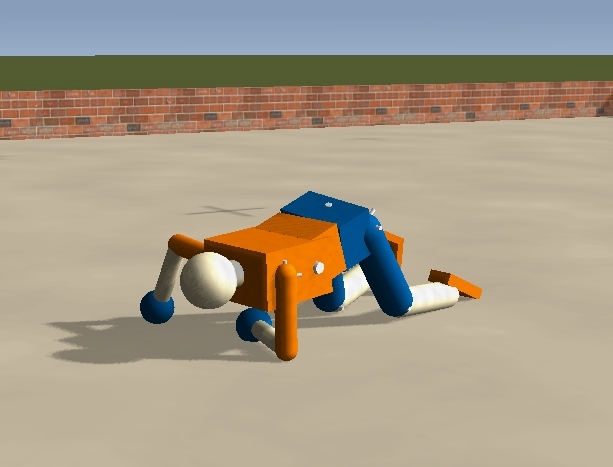}\\[1px] \includegraphics[height=1.68cm]{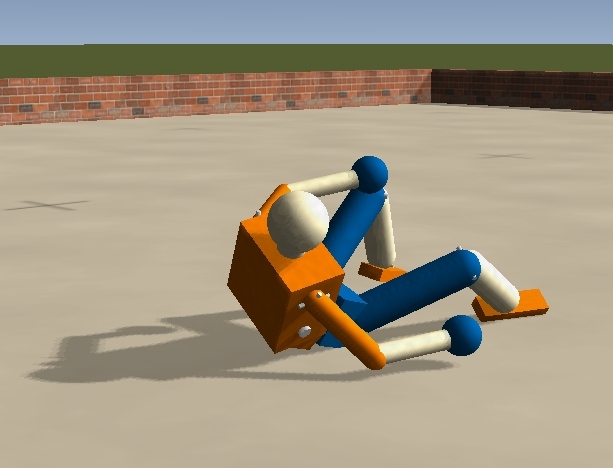} \end{minipage} &
\includegraphics[height=3.4cm]{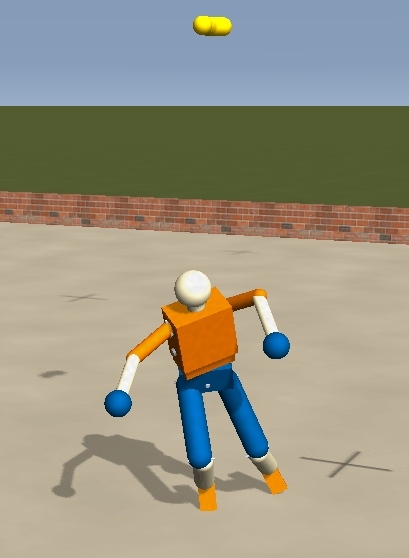} &
\includegraphics[height=3.4cm]{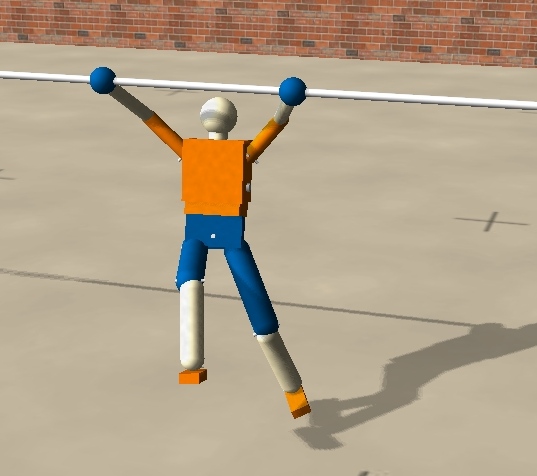} &
\begin{minipage}[b]{1.7cm} \includegraphics[height=1.68cm]{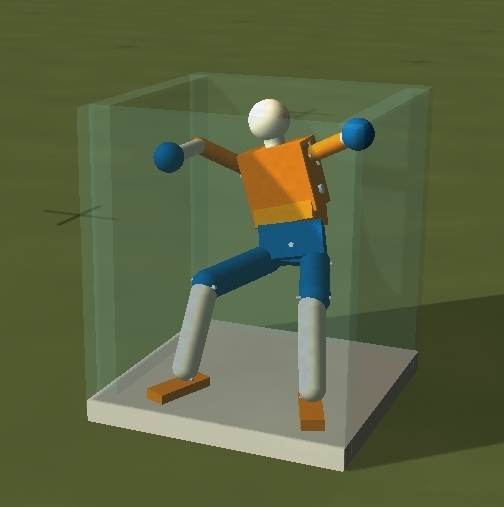}\\[1px] \includegraphics[height=1.68cm]{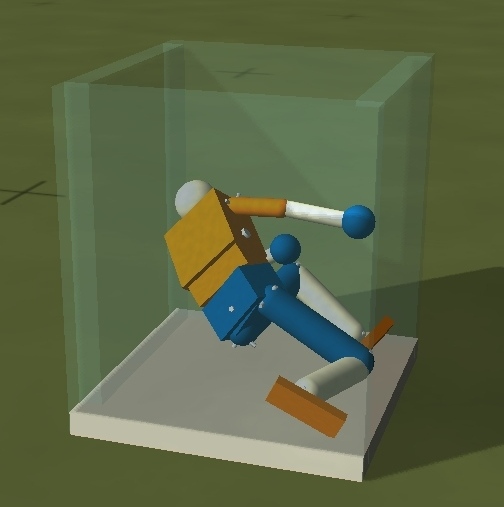} \end{minipage}
\end{tabular}
\vspace*{0.3em}
\caption{{\bf The \Humanoid{} robot in four different scenarios.} (\infig{A}) Normal
environment with flat ground. (\infig{B}) The robot is hanging at a bungee
like spring. (\infig{C}) The robot is attached to a high bar. (\infig{D})
Robot is fallen into a narrow pit.}%
\label{fig:humanoid:scenarios}%
\end{figure}

\begin{figure}
\centering  \includegraphics[scale=0.8]{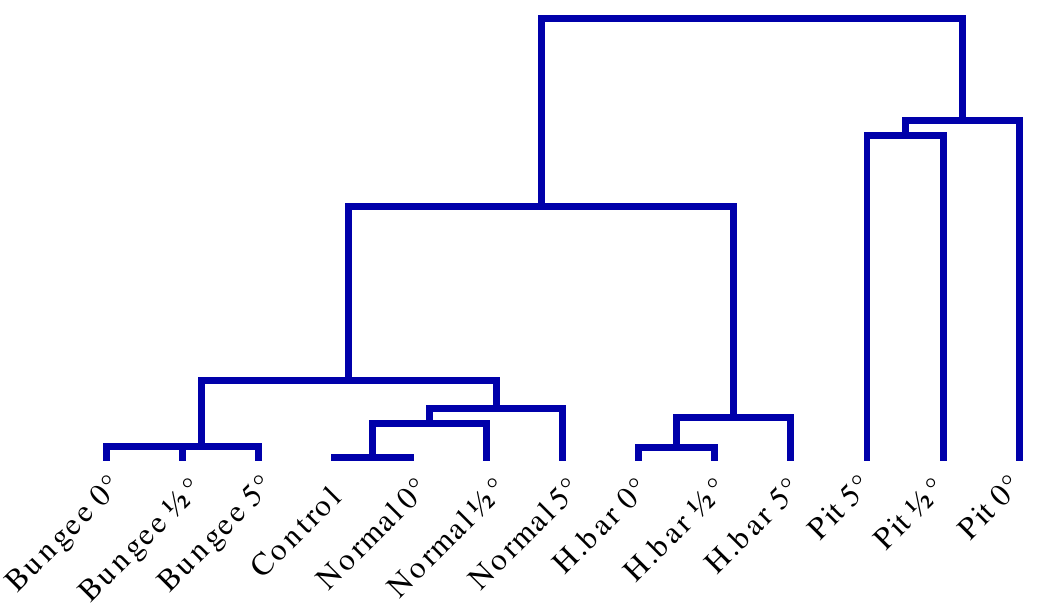}
\caption{{\bf Parameter similarity for the behavior in different environments (\fig{fig:humanoid:scenarios}).}
Plotted is the results of a hierarchical clustering based on the difference
between the parameters in each of the simulations (averaged over time).
For each of the four environments there are three initial poses: $0^\circ$ (straight upright), $0.5^\circ$ and $5^\circ$ slanted to the front.
The parameters for runs in the same environment are clustered together. This
supports the observation that the embodiment plays an essential role in the
generation of behavior. More importantly the physical conditions are reflected
in the parameters and are thus internalized. We used the squared norm of the
difference of the absolute values of the matrix elements. The absolute values
were used because a common structure in the parameters are rotation matrices
and there the same qualitative behavior is obtained with inverted signs.
Parameters: $\varepsilon=0.001$ ($\eta=0.005$)}%
\label{fig:humanoid:dendrogram}%
\end{figure}

That this happens indeed is demonstrated by the videos \suppl{S2, S3, S4 and S5}.
However, we
want a more objective quantity to assess the relation between body and
behavior. We provide two different measures for that purpose.
One idea is to use the parameter constellation of the controller
itself for characterizing the behavior---different behaviors should reflect in
characteristic parameter configurations of the controller. In order to study
this idea, we place the robot in different scenarios, see
\fig{fig:humanoid:scenarios}, always starting with the same initial parameter
configuration (using the result of a preparatory learning phase in the bungee
setting), letting the robot move independently for 40\,min physical
time. Without any additional noise, the dynamics is deterministic so that
variations are introduced by starting the robot in different poses, \ie in a
straight upright position and in slightly tilted poses ($0.5^{\circ}$ and $5^{\circ}$ slanted to the front). We then compared the parameter values of the controller matrix
$C$ at each second (1\,s) for all simulations and calculated a hierarchical clustering reflecting the differences between
the $C$ matrices. \Fig{fig:humanoid:dendrogram} shows the resulting dendrogram.

Obviously, there is a distinct grouping of the $C$ matrices according to the environment the robot is
in and the behaviors developing in the respective situation. Distances between the groups are different, the
most pronounced group corresponding to the behavior in the pit situation. This seems plausible since the
constraints are most distinctive here, driving the robot to behaviors that are markedly  different from
the situation with the bungee setting, say, where all joints (extremities, hip, back) can move much more freely.
There is a second pronounced group---the robot clinging to the high bar---whereas the distances between
the $C$ matrices controlling the robot lying on the ground and hanging at the bungee rope is less pronounced.
However, by visual inspection the emerging behaviors in the two latter
situations appear quite different (compare videos \suppl{S2 and S3})---a finding that is not so clear in the matrix distance method.


\begin{figure}
\centering  \includegraphics[scale=1]{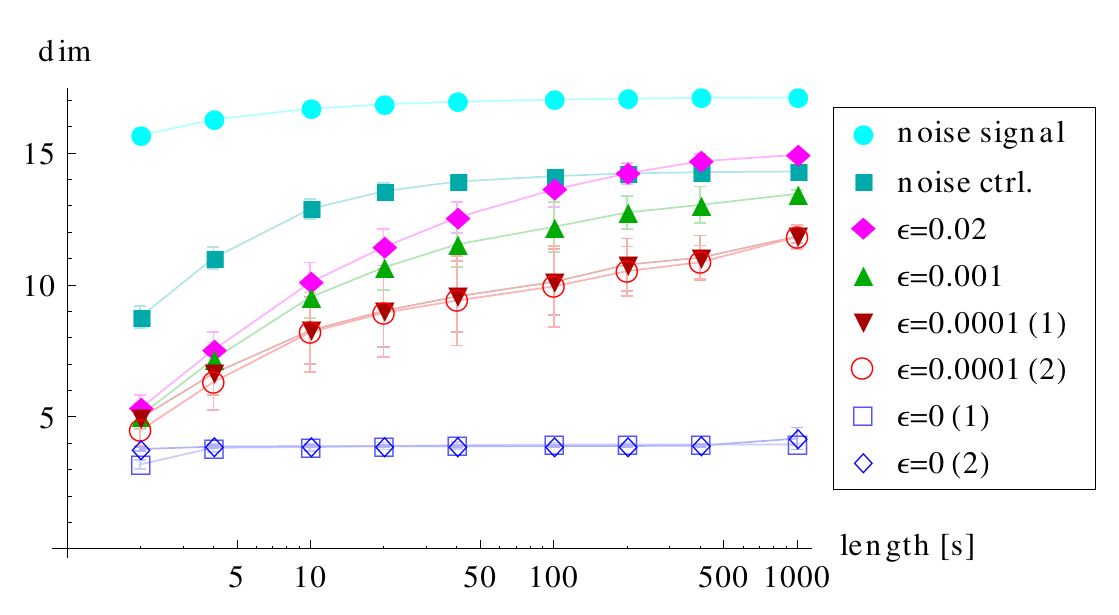}
\caption{{\bf Dimensionality of behavior on different time scales.} \Humanoid{}
robot in bungee setup running 40\,min with different control settings. The sensor
data is partitioned into chunks of a fixed  length, the graph depicting the effective dimension
over the length of the chunks for different settings.
In order to test the method we start with a uniformly distributed noise signal for motor commands
(``noise signal''). As expected the observed dimension is maximal.
The sensor values produced by that random controller show a lower dimension (``noise ctrl.'')
as is expected due to the low pass filtering property of the mechanical system.
All other cases are with the TiPI maximization controller with different update rates $\varepsilon$.
In particular, the comparison with the $\varepsilon=0$ case demonstrates that the exploration dynamics produces
 more complex behaviors than any fixed controller. }%
\label{fig:humanoid:dimensions}%
\end{figure}

\begin{figure}
\centering
\includegraphics[scale=0.95]{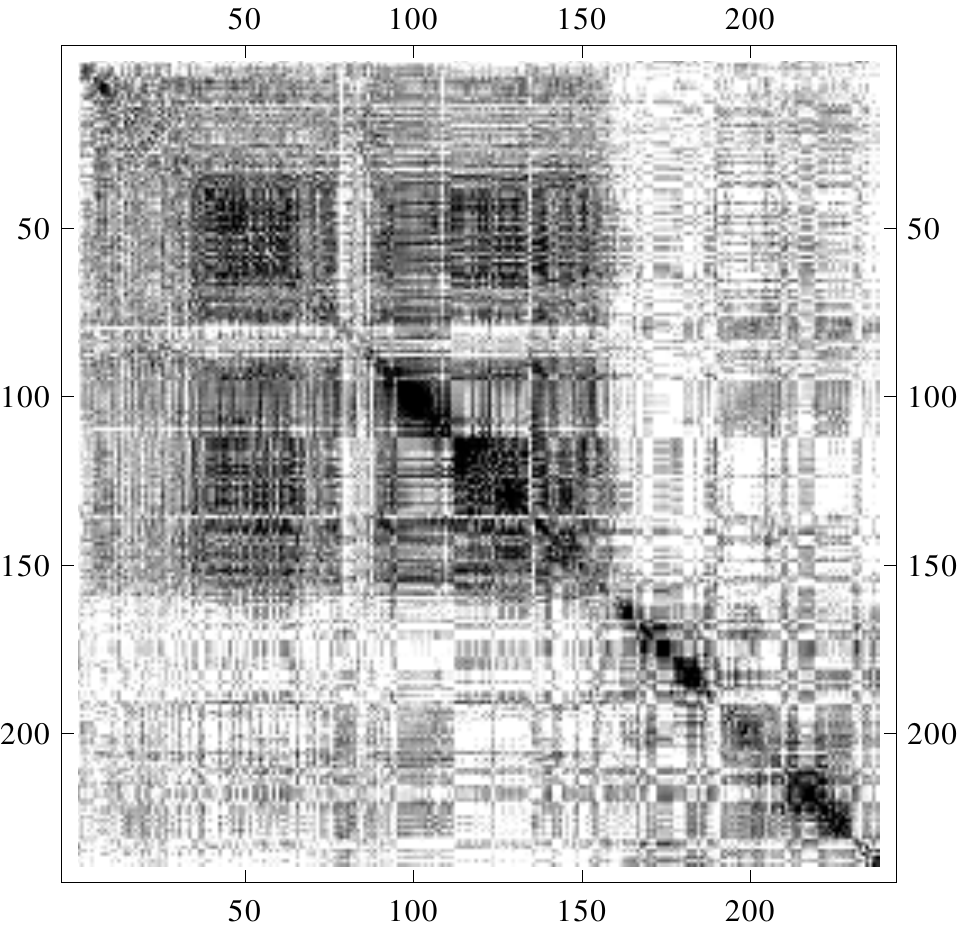}\quad
\includegraphics[scale=0.95,viewport=0 -25 42 200]{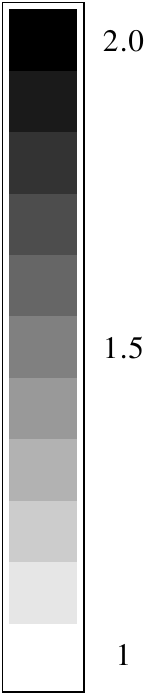}
\caption{{\bf Behavioral changes with time.} Pairwise distances of chunks with
length 10\,s. Distance is defined as the length of the vector of maximal
projections of the first 6 principal components. }%
\label{fig:humanoid:behavior:change}%
\end{figure}

In order to get an additional measure we start from the idea that the TiPI maximization method
produces a series of behaviors that are qualified by a high dynamical complexity generated in a controlled way. The latter point
means that the dimensionality of the time series of the sensor values is much less than that of the
mechanical system -- if the behavior  of the robot is well controlled (think of a walking  pattern)
a few master observables will be sufficient to describe the dynamics of the mechanical system.
We have tried different methods from dynamical system theory for finding the effective
dimension of that time series without much success. The reason was found to be in the strongly nonstationary
nature of the compound dynamics (system plus exploration dynamics) making low dimensional behaviors
to emerge and disappear in a rapid sequence. So, in the long run the full space of the dynamical system is
visited so that globally a seemingly high dimensional behavior is observed.

In order to cope with this nonstationary characteristic, we developed a different method, splitting the whole time series
into chunks and using an elementary principal component analysis (PCA) in order to define the effective dimension in
each chunk: on each chunk a PCA is
performed and the number of principal components required to capture $95\%$ of
the data's variance is plotted (mean and standard deviation for all chunks of
the same length). In order to avoid discretization
artifacts we linearly interpolate the required number of components
to obtain a real number.

The results presented in  \fig{fig:humanoid:dimensions} corroborate
the above hypothesis on the dimensionality of the behaviors. In particular, we
observe the increase of the effective dimension if the chunk length is increasing,
 mixing different low dimensional behaviors.
The latter point is made even more obvious in  \fig{fig:humanoid:behavior:change}
depicting the overlap between the behaviors in chunks at different times.
This overlap is large if the behaviors are essentially the same and small
 if the behavior has changed in the time span between the chunks.
As the figure demonstrates, the overlap is indeed large for short time spans,
 but behaviors can reemerge after some time.
Altogether, the results demonstrate that our TiPI maximization method
effectively explores the behavior space of high-dimensional robotic systems by exciting their low-dimensional
modes, avoiding in this way the curse of dimensionality.

\section{Discussion}

Can a robot develop its skills completely on its own, driven by the sole
objective to gain more and more information about its body and its interaction
with the world? This question raises immediately further issues such as (i) what is
the relevant information for the robot and (ii) how can one find a convenient
update rule that realizes the gradient ascent on this information measure.
We have studied the predictive information of the stream of sensor
values as a tentative answer to the first question and, based on that, could
give exact answers to the second question for simple cases.
Earlier work was restricted to linear systems~\citep{AyDerBernigauProkopenko2012}.
In order to be applicable to actual robotic systems we
 extend it to the case of nonlinear controllers and to nonstationary processes leading to
 a new measure called TiPI (time-local predictive information).
Using several approximations we have been still able to obtained analytical results.
In this way we derived an explicit exploration dynamics for the controller parameters based
 on an information maximization principle, namely by maximizing the TiPI using gradient ascent.
For neural networks the gradient yields a fast synaptic dynamics
 which is essentially local in nature.
Interestingly the TiPI landscape (on which the gradient is calculated)
 continuously changes its shape due to the
general destabilization of the system dynamics inherent in maximizing the TiPI.
For instance if the system dynamics is in an attractor,
 increasing the TiPI destabilizes the attractor until it may disappear altogether
 with a complete restructuring of the TiPI landscape.
This is another reason why nonstationary processes have to be handled and
 why no convergence of the parameter dynamics is desired.

We studied a one-dimensional hysteresis system in order to work out the
 consequences of the nonstationary.
The parameter dynamics leads to a slightly supercritical regime and additionally
 a self-induced hysteresis oscillation emerges.
This is a useful new property as shown in the experiment with the \Armband{} robot,
 a high-dimensional robot with a complicated dynamics.
Despite the highly decentralized control---each joint is controlled individually---the
 robot develops coherent and global pattern of behavior.
This is enabled by the continuous adaptation and spontaneous mutual cooperation of the individual controllers
 (hysteresis elements).
We find the effect to be very robust against the speed of the exploration dynamics.
Interestingly in the one-dimensional case the update formulas are
 independent of white noise and we can obtain an exploration dynamics in a fully deterministic system.

The new theoretical basis also allows for controlling complex high-dimensional robotic systems.
This is demonstrated by a series of experiments with the \Humanoid{} robot,
 now jointly controlled by a single high-dimensional controller.
Given that there is no externally defined goal for the behavior
development, will the robot develop a high behavioral variety depending on its
physics and the environment it is dynamically embedded into?
Our results support a positive answer to this question.
We quantify the dimensionality and temporal structure
of the behavior and find a succession of low-dimensional modes that
 increasingly explore the behavior space.
Furthermore we show that environmental factors influence the internal as well as behavioral development.
Without additional noise, the deterministic dynamics leads to an individual
 development which depends decisively on the particular experiences made during the lifetime.


The exploration dynamics can be viewed as a self-directed search process,
 where the directions to explore are created from the dynamics of the system itself.
Without a random component the changes of the parameters
 are deterministically given as a function of the sensor values and internal parameters
 in a certain time window.
For an embodied system this means in particular that constraints, responses and
 current knowledge of the dynamical interaction with the environment can directly be used to
 advance further exploration.
Randomness is replaced with spontaneity which we demonstrate to restrict the
 search space automatically to the physically relevant dimensions.
Its effectiveness is shown in the \Humanoid{} experiments
 and we argue that this is a promising way to avoid the curse of dimensionality.


What is the relation of the parameter dynamics described here
 to other work on maximizing information quantities  in neural systems?
Maximizing the mutual information between input and output of a neuron, known as InfoMax,
 yields a very similar parameter dynamics~\cite{Bell95infomax}. Interestingly,
 when applied to a feed-forward network an independent component analysis can be performed.
Also similar rules have been obtained in~\citet{Triesch05} where
 the entropy of the output of a neuron was maximized under the condition of
a fixed average output firing-rate~\citet{Triesch05}.
The resulting dynamics is called intrinsic plasticity as it acts on the membrane
 instead of on the synaptic level and it was shown to result in the
 emergence of complex dynamical phenomena~\citep{Triesch06a,Triesch06b,Triesch07,Triesch11}.
In~\citet{MarkovicGros10,MarkovicGros12} a related dynamics is obtained at the
synaptic level of a feedback circuit realized by an autaptic (self) connection.
In a recurrent network of such neurons it was shown that any finite update
rate ($\varepsilon$ in our case) destroys all attractors,
 leading to intermittently bursting behavior and self-organized chaos.

Our work differs in two aspects. On the one hand,
we use the information theoretical principle at
the behavioral level of the whole system by maximizing the TiPI on the full sensorimotor loop,
 whereas they use it at the neuronal level.
Nevertheless we manage to root the information paradigm back to the level
 of the synaptic dynamics of the involved neurons. On the other hand, as a
direct consequence of that approach, there is no need to specify the average
output activity of the neurons. Instead the latter is self-regulating by the
closed loop setting.
Independent of the specific realization, the general message is that these
self-regulating neurons realize a specific working regime where they are both
active and sensitive to influences of their environment. If embedded into
a feedback setting many interesting phenomena are produced.
Instead of studying them in internal (inside the ``brain'') recurrences,
 we embed such neurons into a feedback loop with complex physical systems where
 the self-active and highly responsive nature of these neurons produces
 similar phenomena at the behavioral level.

In the current form, our approach is limited to the control
 of robots where the sensorimotor dynamics can be, in its essence, modeled by a simple
 feed-forward neural network.
The parameter dynamics can also be calculated for more complex controllers, such as recurrent networks,
 which remains for future work.
In this study only proprioceptive sensors measuring joint angles have been used. However, our newest
 experiences have shown that also other sensors \eg current sensors,
 acceleration sensor or velocity sensors can be successfully integrated.

To conclude, information theory is a powerful tool to express principles to drive autonomous systems because it
 is domain invariant and allows for an intuitive interpretation.
We present for the first time, to our knowledge, a method linking
 information theoretic quantities on the behavioral level (sensor values) to explicit dynamical rules
 on the internal level (synaptic weights) in a systematic way.
This opens new horizons for the applicability of information theory to the sensorimotor
 loop and autonomous systems.

\section*{Acknowledgments}
The project was supported by the DFG (SPP 1527).

\bibliographystyle{abbrvnat}

\appendix

\section{Estimating the TiPI\label{sec:error:propagation}}
As stated above we consider the TiPI on the process of error propagations
 because it allows us to derive explicit expressions.
Thus we start with the definition of the error propagation
 to  derive \eqn{eqn:PI:explicit10} and provide further insights.

As a first step, using the notion of an orbit of the dynamical system defined by
the map $\psi:\Real^{n}\rightarrow \Real^{n}$ we define a sequence of states
$\hat{s}_{t^{\prime}}\in \Real^{n}$
\begin{equation}
\hat{s}_{t^{\prime}}=\psi^{t^{\prime}-\left(  t-\tau\right)  }\left(
s_{t-\tau}\right)  \label{eqn:errorforward20}%
\end{equation}
for any time $t^{\prime}$ within the time window $t-\tau\leq t^{\prime}\leq t$ starting from
state $\hat{s}_{t-\tau}=s_{t-\tau}$.
$\psi^k(s)$ denotes the $k$-fold iteration of the map $\psi$ with $\psi^{(0)}\left(s\right)  =s$.
We can consider $\hat{s}_{t^{\prime}}$ as the predicted state over $t^{\prime}-\left(  t-\tau\right)$
time steps.
In particular, the prediction over $\tau$ steps is $\hat{s}_{t}=\psi^{\tau}\left(s_{t-\tau}\right)$.

The error propagation can now be defined as the difference
\begin{equation}
\delta s_{t^{\prime}}=s_{t^{\prime}}-\hat{s}_{t^{\prime}}
\label{eqn:errorforward22a}%
\end{equation}
between the true state $s_{t^{\prime}}$, \eqn{eqn:estimat:PI:psi}, and the state $\hat{s}_{t^{\prime}}$
obtained by the deterministic dynamics ($\psi$), see \fig{fig:timewindow-errorprop}.
The dynamics of the $\delta s_{t^{\prime}}$ obeys the rule\footnote{Proof: Using $\hat
{s}_{t^{\prime}}=\psi\left(  \hat{s}_{t^{\prime}-1}\right)  $ we write
\begin{align*}
\delta s_{t^{\prime}}  &  =s_{t^{\prime}}-\hat{s}_{t^{\prime}}=\psi\left(
s_{t^{\prime}-1}\right)  +\xi_{t^{\prime}}-\psi\left(  \hat{s}_{t^{\prime}-1}\right)\\
&  =\psi\left(  \hat{s}_{t^{\prime}-1}+\delta s_{t^{\prime}-1}\right)
-\psi\left(  \hat{s}_{t^{\prime}-1}\right)  +\xi_{t^{\prime}}\\
&  =L\left(  s_{t^{\prime}-1}\right)  \delta s_{t^{\prime}-1}+\xi_{t^{\prime}} + O(\|\xi\|^2)%
\end{align*}}
\begin{equation}
\delta s_{t^{\prime}}=L\left(  s_{t^{\prime}-1}\right)  \delta s_{t^{\prime
}-1}+\xi_{t^{\prime}} + O(\|\xi_t\|^2) \label{eqn:errorforward25}%
\end{equation}
with starting state $\delta s_{t-\tau}=0$ and $L(s)$ denoting the Jacobian matrix of $\psi$.
In the following we will use this approximation which is arbitrary good for infinitesimally small noise.
Note that this dynamics corresponds to that
of a linear system\footnote{In a linear system, $L$ is independent of the
state. In this case $\hat{s}_{t^{\prime}}=L\hat{s}_{t^{\prime}-1}$ such that
the dynamical evolution of $\delta s$ and $s$ are the same.}, however with state
dependent dynamical operator $L$.

As a remark, in the case of finite noise, we can obtain a related exact rule by using the mean
value theorem of differential calculus stating that under mild restrictions
one can find a state $\tilde{s}_{t^{\prime}}\in\lbrack\hat{s}_{t^{\prime}%
},s_{t^{\prime}}]$ so that
\begin{equation}
\delta s_{t^{\prime}}=L\left(  \tilde{s}_{t^{\prime}-1}\right)  \delta
s_{t^{\prime}-1}+\xi_{t^{\prime}} \label{eqn:errorforward25a}%
\end{equation}
yields the exact dynamics of the multi-step prediction error $\delta s_{t}$.

The interesting point now is that $\Itau\left(  S_{t}:S_{t-1}\right)$ \eqnp{eqn:TiPI} is equal to that of the process defined by the error
propagation dynamics\footnote{Consider two random vectors $S$ and $S^{\prime}$
together with the shifted vectors $U=S+a$ and $U^{\prime}=S^{\prime}%
+a^{\prime}$. Using that the probability distribution functions (pdf)
$p_{S}\left(  s\right)  $ and $p_{U}\left(  u\right)  $ obey $p_{U}\left(
u\right)  =p_{U}\left(  s+a\right)  =p_{S}\left(  s\right)  $ one obtains
$H\left(  S\right)  =H\left(  U\right)  $. Analogously, the joint pdf's obey
$p_{UU^{\prime}}\left(  u,u^{\prime}\right)  =p_{UU}\left(  s+a,s^{\prime
}+a^{\prime}\right)  =p_{SS^{\prime}}\left(  s,s^{\prime}\right)  $ so that
$H\left(  S^{\prime}|S\right)  =H\left(  U^{\prime}|U\right)  $.}\!\!, \ie
\begin{equation}
\Itau\left(  S_{t}:S_{t-1}\right)  =\Itau\left(  \delta S_{t}:\delta S_{t-1}\right)\period
\label{eqn:Idelta_gleich_IS}
\end{equation}
This result is central for the following arguments---we will make use of the
fact that the dynamics \eqn{eqn:errorforward25} is more easily treated to
obtain explicit estimates for the TiPI and its gradient.

\subsubsection*{Explicit expressions }

By iterating \eqn{eqn:errorforward25} we obtain an explicit expression for
$\delta s_{t}$ (using here and in the following $L\left(  t^{\prime}\right)  $
for $L\left(  s\,_{t^{\prime}}\right)  $)%
\begin{equation}
\delta s_{t}=\sum_{k=0}^{\tau-1}L^{\left(  k\right)  }\left(  t-1\right)
\xi_{t-k} \label{eqn:Nonlincontro10}%
\end{equation}
with
\begin{equation}
L^{\left(  k\right)  }\left(  t-1\right)  =L\left(  t-1\right)  \cdots
L\left(  t-k\right)  \text{, and }L^{\left(  0\right)  }=\mathbf{I}\text{ }
\label{eqn:Nonlincontro12}%
\end{equation}
for any $t$. In general it is very complicated to obtain
 the entropy of $\delta S_{t}$ in realistic situations with high dimensional
physical systems. Therefore we will base the further considerations on a
convenient estimate of the latter. With white Gaussian noise, the process
$\delta S_{t}$ is Gaussian as well, \ie$\delta S_{t}\sim\mathcal{N}\left(
0,\Sigma_{t}\right)  $ (it is a linear combination of independent Gaussians),
so that the entropy is given in terms of the covariance matrix $\Sigma_{t}$ of
the random vector $\delta S_{t}$ as~\citep{CoverThomas06}
\begin{equation}
\Htau\left(  \delta S_{t}\right)  =\frac{1}{2}\ln\left\vert \Sigma_{t}\right\vert
+\frac{n}{2}\ln2\pi e \label{eqn:Nonlincontro20}%
\end{equation}
$\left\vert A\right\vert $ denoting the determinant of a square matrix $A$ and
\begin{equation}
\Sigma_{t}=\left\langle \delta S_{t}\delta S_{t}^{\T}\right\rangle =\int
p\left(  \delta s_{t}\right)  \delta s_{t}\delta s_{t}^{\T}\dint \delta s_{t}
\label{eqn:Nonlincontro20a}%
\end{equation}
is the covariance matrix of $\delta S_{t}$ and $p\left(  \delta s_{t}\right)$
is the probability density distribution of the random variable $\delta
S_{t}$. Using \eqn{eqn:Nonlincontro10}, explicit expressions for $\Sigma$ can
readily be obtained, see \eqn{eqn:PI explicit20} below.

By the same arguments, the conditional entropy is defined, using
\eqn{eqn:estimat:PI10}, as
\begin{equation}
\Htau\left(  \delta S_{t}|\delta S_{t-1}\right)  =\Htau\left(  \Xi_{t}\right)
=\frac{1}{2}\ln\left\vert D_{t}\right\vert +\frac{n}{2}\ln2\pi e
\label{MDP210}%
\end{equation}
with
\begin{equation}
D_{t}=\left\langle \Xi_{t}\Xi_{t}^{\T}\right\rangle =\int p\left(  \xi
_{t}\right)  \xi_{t}\xi_{t}^{\T}\dint\xi_{t} \label{eqn:Nonlincontro20b}%
\end{equation}
where $\Xi$ denotes the process of the noise with $p\left(  \xi\right)  $ being the
 probability density function of $\Xi\sim\mathcal{N}\left(0,D_{t}\right)$.
Thus we obtain the estimate of the TiPI as
\begin{equation}
\Itau\left(  \delta S_{t}:\delta S_{t-1}\right)  =\frac{1}{2}\ln\left\vert
\Sigma_{t}\right\vert -\frac{1}{2}\ln\left\vert D_{t}\right\vert
\label{eqn:PI:explicit10:app}%
\end{equation}
which is the entropy of the state $\delta s$ minus that of the noise.

\subsubsection*{White noise}

Explicit expressions revealing more details of the theory are obtained for the
case of white noise, meaning $\left\langle \xi_{t}\xi_{t^{\prime}}^{\T}\right\rangle
=\mathbf{0}$ if $t\neq t^{\prime}$, so that using \eqn{eqn:Nonlincontro10} in
\eqn{eqn:Nonlincontro20a} yields
\begin{equation}
\Sigma=\sum_{k=0}^{\tau-1}L^{\left(  k\right)  }D\left(  L^{\left(  k\right)
}\right)^{\T}\period \label{eqn:PI explicit20}%
\end{equation}
In particular, in the case of $\tau=2$, the shortest nontrivial time window,
we find
\[
\Sigma=D+LDL^{\T}.%
\]
It is also useful to introduce the transformed dynamical operator%
\footnote{This corresponds to using a so-called whitening transformation on
the state dynamics, replacing in \eqn{eqn:errorforward25a} the state vector
$\delta s$ by a new vector $\delta x=\sqrt{D^{-1}}\delta s$ so that the
covariance matrix of the noise in the $\delta x$ dynamics is just the unit
matrix.} $\hat{L}=\sqrt{D^{-1}}L\sqrt{D}$ which leads to  $\Sigma=\sum_{k=0}^{\tau
-1}\sqrt{D}\hat{L}^{\left(  k\right)  }\left(  \hat{L}^{\left(  k\right)
}\right)  ^{\T}\sqrt{D}$ and (using $\left\vert \sqrt{D}M\sqrt{D}\right\vert
=\left\vert MD\right\vert =$ $\left\vert M\right\vert \left\vert D\right\vert
$)%
\begin{equation}
\Itau\left(  \delta S_{t}:\delta S_{t-1}\right)  =\frac{1}{2}\ln\left\vert
\sum_{k=0}^{\tau-1}\hat{L}^{\left(  k\right)  }\left(  \hat{L}^{\left(
k\right)  }\right)  ^{\T}\right\vert \period \label{eqn:PI explicittau2}%
\end{equation}
Interestingly, the $\hat{L}$ operators also exist if the overall noise
strength $\lambda=\left\Vert \xi\right\Vert $ goes to zero, so that $\Itau$ stays
finite\footnote{Introducing $\hat{D}=\lambda^{-2}D$ where $\hat{D}$ stays
finite with $\lambda\rightarrow0$,
we have $\hat{L}=\sqrt{\hat{D}^{-1}}L\sqrt{\hat{D}} =
\sqrt{D}L\sqrt{D}$ since $\lambda$ cancels out.}
although the defining entropies, conditioned on the state
$s_{t-\tau}$, are equal to zero in the deterministic system.

\subsubsection*{The linear case }

For linear systems explicit expressions for the PI were obtained in
\citet{AyDerBernigauProkopenko2012}.
In this case $L$ is not dependent on the state
$s_{t}$ of the system so that $L^{\left(  k\right)  }=L^{k}$ in
\eqn{eqn:Nonlincontro12}. Using \eqn{eqn:PI explicit20}, with $\tau
\rightarrow\infty$, we reobtain the results\footnote{Note that all eigenvalues
of the Jacobi matrix $L$ must be less than one by absolute value so that the
limes will exist. This requirement also guarantees that the conditioning on
$s_{t-\tau}$ looses its influence for $\tau\rightarrow\infty$.} of
\citet{AyDerBernigauProkopenko2012}.
Under the additional assumption that $L$ is a normal matrix and the noise is isotropic the explicit expression
 $\Sigma = \left( \mathbb{I} - L L^\top\right)$ was obtained.

\section{Explicit gradient step\label{sec:LearningRules:General}}
In order to derive the general gradient step on the TiPI based on \eqn{eqn:LearningRulesIntro10a}
 we need to calculate the derivative $\frac{\partial}{\partial\theta}\ln\left|\Sigma_t\right|$.
Considering any (square) matrix $M$ depending on a single parameter
$\theta_k$ of the set $\theta$ we have%
\footnote{We write $\frac{1}{M}$ for $M^{-1}$ here and in the following.} (see
for example~\citet{Magnus1988})
\[
\frac{\partial}{\partial M}\ln\left\vert M\right\vert =\frac{1}{M^{\T}}%
\]
and
\[
\frac{\partial}{\partial\theta_k}\ln\left\vert M\right\vert =\sum_{ij}%
M_{ji}^{-1}\frac{\partial M_{ij}}{\partial\theta_k}=\Tr\left(  \left(
M^{-1}\right)  ^{\T}\frac{\partial M}{\partial\theta_k}\right)
\]
so that, using $\Sigma=\Sigma^{\T}=\left\langle \delta s\delta
s^{\T}\right\rangle $ and omitting the time index
\begin{equation}
\frac{\partial}{\partial\theta_k}\ln\left\vert \Sigma\right\vert =\Tr\left(
\frac{1}{\Sigma}\frac{\partial}{\partial\theta_k}\left\langle \delta s\delta
s^{\T}\right\rangle \right)\period  \label{eqn:LearningRulesIntro20a}%
\end{equation}

By using the cyclic invariance of the trace
we obtain from \eqn{eqn:LearningRulesIntro20a}
\begin{equation}
\frac{\partial}{\partial\theta}\ln\left|\Sigma_t\right|=\left\langle \delta s_{t}^{\T}\Sigma
^{-1}\frac{\partial}{\partial\theta}\delta s_{t}\right\rangle
\label{eqn:AppLearningRuleGeneral10}%
\end{equation}
now valid for the entire set of parameters $\theta$.
By \eqn{eqn:errorforward25a} we obtain (ignoring the dependence of $\xi$ on
the parameter)%
\[
\frac{\partial}{\partial\theta}\delta s_{t^{\prime}}=\frac{\partial L\left(
t^{\prime}-1\right)  }{\partial\theta}\delta s_{t^{\prime}-1}+L\left(
t^{\prime}-1\right)  \frac{\partial}{\partial\theta}\delta s_{t^{\prime}-1}%
\]
so that by iteration %
\[
\frac{\partial}{\partial\theta}\delta s_{t}=\sum_{l=1}^{\tau-1}L^{\left(
l-1\right)  }\left(  t-1\right)  \frac{\partial L\left(  t-l\right)
}{\partial\theta}\delta s_{t-l}%
\]
where $L^{\left(  k\right)  }\left(  t-1\right)  $ is given in
\eqn{eqn:Nonlincontro12}. Using $a^{\T}Wb=\left(  W^{\T}a\right)  ^{\T}b$, we
write
\begin{equation}
\frac{\partial}{\partial\theta}\ln\left|\Sigma_t\right|=\sum_{l=1}^{\tau-1}\left\langle \delta
u_{t-l+1}^{\T}\frac{\partial L\left(  t-l\right)  }{\partial\theta}\delta
s_{t-l}\right\rangle \label{eqn:AppLearningRuleGeneral10a}%
\end{equation}
where ($\Sigma$ is symmetric)
\begin{equation}
\delta u_{t-l+1}=\left(  L^{\left(  l-1\right)  }\left(  t-1\right)  \right)
^{\T}\Sigma_{t}^{-1}\delta s_{t} \label{eqn:AppLearningRuleGeneral10u}%
\end{equation}
Stipulating the self-averaging property of the stochastic gradient,
see section One-shot gradients for details, we realize the 
update rule as
\begin{equation}
\Delta\theta=\varepsilon\sum_{l=1}^{\tau-1}\delta u_{t-l+1}^{\T}\frac{\partial
L\left(  t-l\right)  }{\partial\theta}\delta s_{t-l}
\label{eqn:AppLearningRuleGeneral10final}%
\end{equation}
Here we see again that $\tau=2$ is the simplest non-trivial case where the sum consists of a single term.


\subsubsection*{Characterizing the parameter dynamics}
In order to better characterize the parameter dynamics, let us consider for
the moment $\Sigma$ at the \rhs of
\eqn{eqn:LearningRulesIntro20a} to be some fixed, positive matrix (not depending
on the parameters $\theta_k$). Then, we can write
\[
\Tr\left(  \frac{1}{\Sigma}\frac{\partial}{\partial\theta_k}\left\langle \delta
s\delta s^{\T}\right\rangle \right)  =\frac{\partial}{\partial\theta_k
}\left\langle \Tr\left(  \frac{1}{\Sigma}\delta s\delta s^{\T}\right)
\right\rangle =\frac{\partial}{\partial\theta_k}\left\langle \delta s^{\T}%
\frac{1}{\Sigma}\delta s\right\rangle
\]
(using the cyclic invariance of the trace in the last step). The update
rule \eqn{eqn:LearningRulesIntro10a} becomes using again the self-averaging
\begin{equation}
\Delta\theta=\varepsilon\frac{\partial}{\partial\theta}\left\Vert \delta
s\right\Vert _{\Sigma}^{2} \label{eqn:LearningRulesIntro30}%
\end{equation}
where
$\left\Vert a\right\Vert _{M}^{2}=a^{\T}M^{-1}a$
defines the length of a vector $a$ in the metric given by $M$ (considered
fixed in the current gradient step).
From \eqn{eqn:LearningRulesIntro30} it becomes obvious that
following the gradient is to increase the norm of $\delta s$ in
the $\Sigma$ metric.

\section{Neural networks---derivation of the update rule\label{SecAppendix:NeuralLearn}}
We derive the parameter dynamics for neural networks \eqn{eqn:NeuralControl:LRa} from
 the general parameter dynamics for the two-step time window given by \eqn{eqn:LearningRulesSpecial24a}.
According to \eqn{eqn:NeuralControlForward10L} we have $L=VG^{\prime}\left(  z\right)  C+T$ with
$z=Cs+h$ and $G'(z)=\mathrm{diag}[g_1^{\prime}(z),\dots,g_m^{\prime}(z)]$.
Putting this into \eqn{eqn:LearningRulesSpecial24a} yields (omitting the time indices)
\begin{align}
\frac{1}{\varepsilon} \Delta C_{ij}&=\delta u^{\T}\frac{\partial L}{\partial C_{ij}}\delta s\nonumber\\
  &= \delta u^{\T}VG^{\prime}\frac{\partial C}{\partial C_{ij}}\delta s+ \delta u^{\T}V\frac{\partial G^{\prime}}{\partial C_{ij}}C \delta s\nonumber\\
  &= \left(G^{\prime}V^{\T}\delta u\right)_{i}\delta s_{j}+ \delta u^{\T}V\frac{\partial G^{\prime}}{\partial C_{ij}}C \delta s\period\label{eqn:Appendix:NeuralLearn:C:start}
\end{align}
The second term remains to be calculated.
Because $G'$ is a diagonal matrix the vectors on both sides of the derivative
 carry the index $i$ such that we get
\begin{equation}
\delta u^{\T}V\frac{\partial G^{\prime}}{\partial C_{ij}}C \delta s
 = \left(\delta u^{\T}V\right)_i g_i^{\prime\prime}(z) s_j \left(C \delta s\right)_i\period
\end{equation}
In the case of $g\left(  z\right)  =\tanh\left(  z\right)  $ we find, using $g_i^{\prime\prime}\left(  z\right)  =-2g_i^{\prime}\left(  z\right)  g_i\left(z\right)$
 and $a=g(z)$
\begin{equation}
\delta u^{\T}V\frac{\partial G^{\prime}}{\partial C_{ij}}C \delta s
 = - 2\left(\delta u^{\T}V G^{\prime} \right)_i \left(C \delta s\right)_i a_i  s_j
 = - \gamma_i a_i s_j \label{eqn:Appendix:NeuralLearn:right}
\end{equation}
with
\begin{align}
\gamma_{i}&=2\left(  C\delta s\right)  _{i}\delta\mu_{i}\\
\delta\mu_{i}&=\left(  G^{\prime}V^{\T}\delta u\right)_{i}\period\label{eqn:Appendix:NeuralLearn:dmu}
\end{align}
The final update rule follows by putting \eqn{eqn:Appendix:NeuralLearn:right}
 and \eqn{eqn:Appendix:NeuralLearn:dmu}
 into \eqn{eqn:Appendix:NeuralLearn:C:start}
\begin{equation}
\Delta C_{ij}=\varepsilon\delta\mu_{i}\delta s_{j}-\varepsilon\gamma_{i}a_{i}s_{j}
\period\label{eqn:Appendix:NeuralLearnFinal}%
\end{equation}

Analogously we obtain the parameter dynamics of $h$ as
\begin{align}
\frac{1}{\varepsilon} \Delta h_{i}&=\delta u^{\T}\frac{\partial L}{\partial h_{i}}\delta s
  = \delta u^{\T}V\frac{\partial G^{\prime}}{\partial h_{i}}C \delta s\nonumber\\
  &= \left(\delta u^{\T}V\right)_i g_i^{\prime\prime}(z) \left(C \delta s\right)_i
  = - \gamma_i a_i \label{eqn:Appendix:NeuralLearnFinal:h} \period
\end{align}

A more compact matrix notation can be obtained by introducing the diagonal matrix $\Gamma$
\[
\Gamma=\mathrm{diag}[\gamma_{1},\cdots,\gamma_{i}]%
\]
and thus (reintroducing the time indices)
\begin{align}
\frac{1}{\varepsilon}\Delta C_{t}  &  =\delta\mu_{t}\delta s_{t-1}%
^{\T}-\Gamma_{t}a_{t}s_{t}^{\T}\komma\label{eqn:NeuralControl:LRaa}\\
\frac{1}{\varepsilon}\Delta h_{t}  &  =-\Gamma_{t}a_{t}%
\period \label{eqn:NeuralControl:LRbb}%
\end{align}

In the case of arbitrary neuron activation functions $g$ we obtain equivalent
formula by defining
\begin{equation}
\gamma_{i}=-\frac{g_{i}^{\prime\prime}}{g_{i}^{\prime}g_{i}}\left(  C\delta
s\right)  _{i}\delta\mu_{i} \label{eqn:Appendix:NeuralLearnFinalG}%
\end{equation}
Note the factor $-\frac{g_{i}^{\prime\prime}}{g_{i}^{\prime}g_{i}}$ is $2$ in the case of $g=\tanh$.

In the derivation of \eqns{eqn:Appendix:NeuralLearn:right}{eqn:Appendix:NeuralLearnFinal:h}
 we ignored the dependence of the state $s$ in $g^{\prime}\left(C s + h\right)$ on the
parameters $C$ and $h$. This dependence can be considered explicitly if the state
is at a fixed point. In that case, a more detailed discussion in~\citet{DerMartius11} (section 6.2)
shows that the effect of the derivative can be condensed into the so-called
\emph{sense} parameter $\alpha$ multiplying $\gamma$.
Thus we replace $\gamma$ as
\begin{equation}
\gamma_{i}\leftarrow\alpha\gamma_{i} \label{eqn:Appendix:Neuralsense}%
\end{equation}
where $\alpha$ is an empirical constant, typically $\alpha\geq1$, by which the
sensitivity of the sensorimotor dynamics to external perturbations can be
regulated. This works also in more general cases like a limit cycle dynamics,
see~\citet{DerMartius11}.

\section{Learning the inverse covariance matrix\label{Sec:SherMorrFormel}}

Note that the covariance matrix given in \eqn{eqn:LearningRulesSpecial24aa}
can be easily obtained by the on-line update rule
\begin{equation}
\Delta\Sigma_{t}=\eta\left(  \delta s_{t}\delta s_{t}^{\T}-\Sigma_{t}\right)
\label{eqn:LearningRulesSpecial26a}%
\end{equation}
or
\begin{equation}
\Sigma_{t+1}=\left(  1-\eta\right)  \Sigma_{t}+\eta\delta s_{t}\delta
s_{t}^{\T} \label{eqn:LearningRulesSpecial26b}%
\end{equation}
realizing a sampling over a restricted period of time. The update rate $\eta$
defines the time horizon $t_{H}\propto\eta^{-1}$ for the averaging. The only
remaining nontrivial operation in that setting is the inversion of the
covariance matrix $\Sigma$. However, this can also be reduced to elementary
operations by using the Sherman-Morrison formula as given by
\[
\left(  A+uv^{\T}\right)  ^{-1}=A^{-1}-\frac{1}{1+v^{\T}A^{-1}u}A^{-1}%
uv^{\T}A^{-1}%
\]
Putting $A=\left(  1-\eta\right)  \Sigma$ and $uv^{\T}=\eta\delta s\delta
s^{\T}$ we get
\[
\left(  \left(  1-\eta\right)  \Sigma_{t}+\eta\delta s_{t}\delta s_{t}%
^{\T}\right)  ^{-1}\!\!=\frac{1}{1-\eta}\Sigma_{t}^{-1}-\frac{\eta}{\left(
1-\eta\right)  ^{2}\left(  1+\frac{\eta}{1-\eta}\delta s_{t}^{\T}\Sigma
_{t}^{-1}\delta s_{t}\right)  }\Sigma_{t}^{-1}\delta s_{t}\delta s_{t}%
^{\T}\Sigma_{t}^{-1}%
\]
and thus
\[
\Sigma_{t+1}^{-1}=\frac{1}{1-\eta}\Sigma_{t}^{-1}-\frac{\beta}{1-\eta}%
\Sigma_{t}^{-1}\delta s_{t}\delta s_{t}^{\T}\Sigma_{t}^{-1}%
\]
where $\beta\in \Real$ is given by
\[
\beta=\frac{\eta}{\left(  1-\eta+\eta\delta s_{t}^{\T}\Sigma_{t}^{-1}\delta
s_{t}\right)  }%
\]
Note that $\delta s_{t}^{\T}\Sigma_{t}^{-1}\delta s_{t}$ featuring in the
denominator of $\beta$ is a scalar so that with $\Sigma_{t}^{-1}$ given there
is no matrix inversion to be done.

If $\Sigma_{t}$ is an $n\times n$ matrix, the cost of getting $\Sigma_{t+1}$
is $O\left(  n^{2}\right)  $. This is very favorable if the dimension of the
sensor space is large. Using the above formula, the only true inversion (of order
$O\left(  n^{3}\right)  $) has to be done just once, when starting the process
(with a convenient initialization of $\Sigma$).

\end{document}